\documentclass[11pt,a4paper]{article}
\usepackage{times,latexsym}
\usepackage{url}
\usepackage[T1]{fontenc}
\usepackage{booktabs}
\usepackage{makecell}

\usepackage{enumitem}
\usepackage{csquotes}
\usepackage{multirow}

\usepackage[acceptedWithA]{tacl2018v2}
\usepackage{graphicx}
\usepackage{scalerel,xparse}

\usepackage{microtype}

\usepackage{xspace,mfirstuc,tabulary}

\newcommand\blfootnote[1]{%
  \begingroup
  \renewcommand\thefootnote{}\footnote{#1}%
  \addtocounter{footnote}{-1}%
  \endgroup
}

\newif\iftaclinstructions
\taclinstructionsfalse 
\iftaclinstructions
\renewcommand{\confidential}{}
\renewcommand{\anonsubtext}{(No author info supplied here, for consistency with
TACL-submission anonymization requirements)}
\newcommand{\instr}
\fi

\iftaclpubformat 

\else

\fi

\NewDocumentCommand\emojigem{}{
    \includegraphics[scale=0.12]{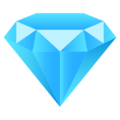}
}

\newcommand{\GEM}{\textsc{GEM}}
\definecolor{cb_purple}{rgb}{0.8, 0.1, 0.1}

\title{The\emojigem\GEM{} Benchmark: \\ Natural Language Generation, its Evaluation and Metrics}

\author{
\begin{minipage}[t]{\textwidth}
\centering
\normalsize
Sebastian~Gehrmann,$^{9,}$\textbf{*}
Tosin~Adewumi,$^{20,21}$
Karmanya~Aggarwal,$^{14}$
Pawan~Sasanka~Ammanamanchi,$^{15}$
Aremu~Anuoluwapo,$^{21,38}$
Antoine~Bosselut,$^{28}$
Khyathi~Raghavi~Chandu,$^{2}$
Miruna~Clinciu,$^{7,11,35}$
Dipanjan~Das,$^{9}$
Kaustubh~D.~Dhole,$^{1}$
Wanyu~Du,$^{42}$
Esin~Durmus,$^{5}$
Ondřej~Dušek,$^{3}$
Chris~Emezue,$^{21,30}$
Varun~Gangal,$^{2}$
Cristina~Garbacea,$^{39}$
Tatsunori~Hashimoto,$^{28}$
Yufang~Hou,$^{13}$
Yacine~Jernite,$^{12}$
Harsh~Jhamtani,$^{2}$
Yangfeng~Ji,$^{42}$
Shailza~Jolly,$^{6,29}$
Mihir Kale,$^{9}$
Dhruv~Kumar,$^{44}$
Faisal~Ladhak,$^{4}$
Aman~Madaan,$^{2}$
Mounica~Maddela,$^{8}$
Khyati~Mahajan,$^{34}$
Saad~Mahamood,$^{32}$
Bodhisattwa~Prasad~Majumder,$^{37}$
Pedro~Henrique~Martins,$^{16}$
Angelina~McMillan-Major,$^{43}$
Simon~Mille,$^{26}$
Emiel~van Miltenburg,$^{31}$
Moin~Nadeem,$^{22}$
Shashi~Narayan,$^{9}$
Vitaly~Nikolaev,$^{9}$
Rubungo~Andre~Niyongabo,$^{21,36}$
Salomey~Osei,$^{19,21}$
Ankur~Parikh,$^{9}$
Laura~Perez-Beltrachini,$^{35}$
Niranjan~Ramesh~Rao,$^{24}$
Vikas~Raunak,$^{23}$
Juan Diego~Rodriguez,$^{41}$
Sashank Santhanam,$^{34}$
João~Sedoc,$^{25}$
Thibault~Sellam,$^{9}$
Samira~Shaikh,$^{34}$
Anastasia~Shimorina,$^{33}$
Marco Antonio~Sobrevilla~Cabezudo,$^{40}$
Hendrik~Strobelt,$^{13}$
Nishant~Subramani,$^{17,21}$
Wei~Xu,$^{8}$
Diyi~Yang,$^{8}$
Akhila~Yerukola,$^{27}$
Jiawei~Zhou$^{10}$ \\

{\footnotesize \normalfont 
$^{1}$Amelia R\&D, New York,
$^{2}$Carnegie Mellon University,
$^{3}$Charles University, Prague,
$^{4}$Columbia University,
$^{5}$Cornell University,
$^{6}$DFKI, Germany
$^{7}$Edinburgh Centre for Robotics,
$^{8}$Georgia Tech,
$^{9}$Google Research,
$^{10}$Harvard University,
$^{11}$Heriot-Watt University,
$^{12}$Hugging Face,
$^{13}$IBM Research,
$^{14}$IIIT Delhi,
$^{15}$IIIT Hyderabad,
$^{16}$Instituto de Telecomunicações,
$^{17}$Intelligent Systems Lab, Intel,
$^{18}$Johns-Hopkins University,
$^{19}$Kwame Nkrumah University of Science and Technology
$^{20}$Luleå University of Technology,
$^{21}$Masakhane, Africa,
$^{22}$Massachusetts Institute of Technology,
$^{23}$Microsoft,
$^{24}$National Institute of Technology Karnataka India,
$^{25}$New York University,
$^{26}$Pompeu Fabra University,
$^{27}$Samsung Research,
$^{28}$Stanford University,
$^{29}$Technical University of Kaiserslautern,
$^{30}$Technical University Munich,
$^{31}$Tilburg University,
$^{32}$trivago,
$^{33}$Université de Lorraine,
$^{34}$University of North Carolina Charlotte,
$^{35}$University of Edinburgh,
$^{36}$University of Electronic Science and Technology of China,
$^{37}$University of California San Diego,
$^{38}$University of Lagos,
$^{39}$University of Michigan Ann Arbor,
$^{40}$University of São Paulo,
$^{41}$University of Texas at Austin,
$^{42}$University of Virginia,
$^{43}$University of Washington,
$^{44}$University of Waterloo
} 
\end{minipage}
}

\date{}

\begin{document}
\maketitle
\blfootnote{\textbf{*} Correspondence to \texttt{gehrmann@google.com}}
\begin{abstract}
    We introduce \GEM, a living benchmark for natural language Generation (NLG), its Evaluation, and Metrics. 
    Measuring progress in NLG relies on a constantly evolving ecosystem of automated metrics, datasets, and human evaluation standards. 
    Due to this moving target, new models often still evaluate on divergent anglo-centric corpora with well-established, but flawed, metrics. 
    This disconnect makes it challenging to identify the limitations of current models and opportunities for progress.
    Addressing this limitation, \GEM{} provides an environment in which models can easily be applied to a wide set of tasks and in which evaluation strategies can be tested. 
    Regular updates to the benchmark will help NLG research become more multilingual and evolve the challenge alongside models. 
    This paper serves as the description of the data for which we are organizing a shared task at our ACL 2021 Workshop and to which we invite the entire NLG community to participate. 
\end{abstract}

\section{Introduction}

Natural language generation is the task to automatically generate understandable texts, typically using a non-linguistic or textual representation of information as input~\citep{reiter2000building}. These texts aim to fulfill an underlying communicative goal (e.g., \textit{to produce a summary of an article}) while remaining faithful to the input information, fluent, grammatical, and natural-looking. An NLG system needs to be robust to shifts in the data distribution and be able to produce text in many different languages. Finally, it is often desired that repeated interactions with the model produce diverse outputs, for example, to explain concepts in multiple ways or to become a more interesting conversational agent. These optimization objectives can often be conflicting~\citep{hashimoto2019unifying} and, as a result, evaluations that focus only on a single aspect may fail to recognize the drawbacks of a particular method. 
To demonstrate this trade-off, consider an improvement on the CNN-DM summarization dataset~\citep{hermann2015teaching,nallapati2016abstractive} measured by the ROUGE-L metric~\citep{lin2004rouge}.
Since ROUGE only tests the extent to which a generated summary has a lexical overlap with a reference summary, it can erroneously produce high scores for fluent, yet meaningless and unfaithful outputs as long as many of the same words are used~\citep{maynez2020faithfulness,gabriel2020go}. Moreover, ROUGE tends to favor systems that produce longer summaries~\citep{sun2019compare}. 
It is thus crucial to carefully assess the progress of NLG toward all of its goals at the same time in ways that evolve alongside the models. 
This is currently not the case; new models are evaluated on different datasets, most of which focus only on the English language~\citep{bender2019benderrule}, and using these flawed metrics. Moreover, while human evaluations of generated texts can provide complementary insights to automatic evaluation~\citep{manning2020human}, it can also lead to contradicting results since studies often omit crucial replication details and assume different definitions of the measured quantities~\citep{howcroft2020twenty}.

\begin{figure}[t]
\centering
\includegraphics[width=\linewidth]{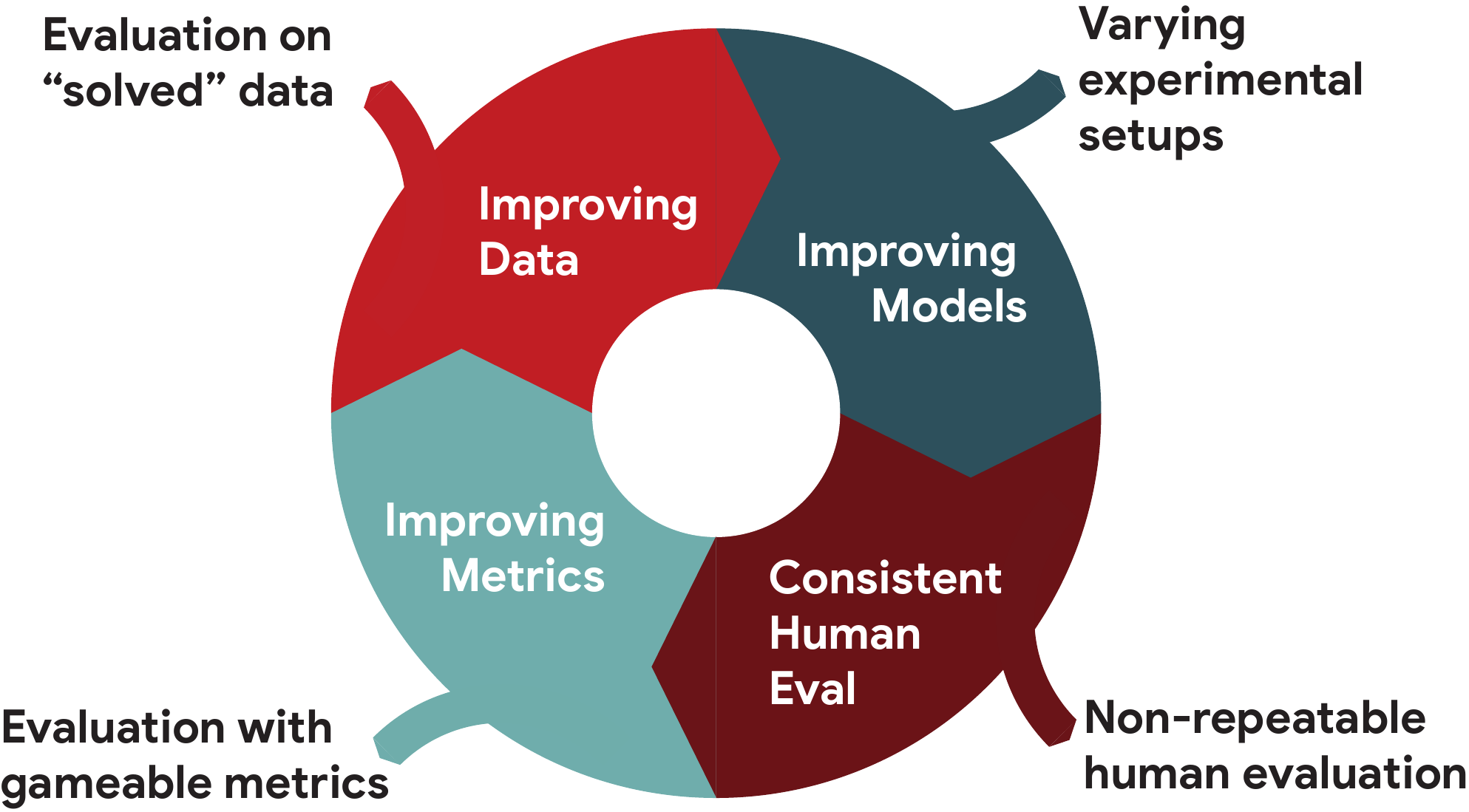}
\caption[The circle of model development]{The opportunities of living benchmarks and pitfalls of evaluation. As models improve, we need consistent evaluations such that models can be compared to each other. This can only happen if we develop robust human evaluation standards and improve our automated metrics. Otherwise, results are challenging to interpret and compare to each other. Finally, as models improve and metrics saturate, we need to evaluate them on more challenging datasets instead of continuing to move sideways on old ones. \GEM{} aims to provide this environment for natural language generation.}
\label{fig:circle}
\end{figure}

We propose a living benchmark called \GEM{} (Generation, Evaluation, and Metrics) that aims to enable research on a wide range of NLG challenges. To avoid the fallacy of encouraging hill climbing on a leaderboard~\citep{linzen2020accelerate}, \GEM{} focuses on an in-depth evaluation of model outputs across human and automatic evaluation that aims to uncover shortcomings and opportunities for progress. As datasets, metrics, and models improve, the benchmark environment will improve as well, replacing ``solved'' tasks with more challenging ones, incorporating newly developed metrics, and addressing discovered flaws in the experimental setup, as demonstrated in Figure~\ref{fig:circle}.
Making all model outputs available under an open-source license will support evaluation research and integrating new metrics will, in turn, help their adoption and increase the robustness of model evaluations. 

The initial set of eleven included datasets is presented in Table~\ref{tab:overview}. They measure specific generation challenges, such as the content selection and planning  (\textit{What to say?}), and the surface realization (\textit{How to say it?})~\citep{reiter2000building,gatt2018survey}. Models need to be capable of paraphrasing, simplification, and others. In addition to those challenges, \GEM{} datasets also differ in their communicative goals, languages, the noisiness of data, and resource availability, to evaluate the consistency of evaluation schemes.
About half of the datasets have multiple references and more than half were post-processed to improve data quality. The sizes range from 5k to 500k data points. \GEM{} features 18 languages across all tasks and two of the datasets do not include English at all.
To be able to properly assess the performance of models in a way robust to the shortcuts a model can take, we additionally introduce ten types of challenging test sets that probe for specific modeling aspects~\citep{perez2017analysing,ribeiro-etal-2020-beyond}. 
To ensure that research with \GEM{} is conducted responsibly, all the datasets are documented in an NLG-specific version of data cards~\citep{bender2018data,gebru2018datasheets} we developed and for which we release a template and guide. Moreover, all submitted models will have an associated data card~\citep{mitchell2019model}.

\setlength{\tabcolsep}{4pt}
\renewcommand{\arraystretch}{0.75}
\begin{table*}[t!]
\small
\begin{tabular}{@{}m{11.9em}m{16.2em}m{5.2em}rm{6em}@{}}
\toprule
\textbf{Dataset} & \textbf{Communicative Goal}  & \centering\arraybackslash \textbf{Language(s)} & \textbf{Size} & \centering\arraybackslash\textbf{Input Type} \\ \toprule
\makecell[bl]{\small{CommonGEN}\\ \small{\citep{lin2019commongen}}} & \small{Produce a likely sentence which mentions all of the source concepts.}  & \centering\arraybackslash\small{en} & \centering\arraybackslash\small{67k} &  \centering\arraybackslash\small{Concept Set} \\ \midrule
\makecell[bl]{\small{Czech Restaurant}\\ \small{\citep{duvsek2019neural}}} & \small{Produce a text expressing the given intent and covering the specified attributes.} & \centering\arraybackslash\small{cs} & \centering\arraybackslash\small{5k} &  \centering\arraybackslash\small{Meaning\quad Representation} \\ \midrule
\makecell[bl]{\small{DART}\\ \small{\citep{radev2020dart}}} & \small{Describe cells in a table, covering all information provided in triples.} & \centering\arraybackslash\small{en} & \small{82k} & \centering\arraybackslash\small{Triple Set} \\ \midrule
\makecell[bl]{\small{E2E clean}\\ \small{\citep{novikova2017e2e}} \\ \small{\citep{duvsek2019semantic}}} & \small{Describe a restaurant, given all and only the attributes specified on the input.} & \centering\arraybackslash\small{en} & \centering\arraybackslash\small{42k}   & \centering\arraybackslash\small{Meaning\quad Representation}  \\ \midrule
\makecell[bl]{\small{MLSum}\\ \small{\citep{scialom2020mlsum}}} & \small{Summarize relevant points within a news article} &  \centering\arraybackslash \small{*de/es} & \centering\arraybackslash\small{*520k}  & \centering\arraybackslash\small{Articles} \\\midrule
\makecell[bl]{\small{Schema-Guided Dialog}\\ \small{\citep{rastogi2019towards}}} & \small{Provide the surface realization for a virtual assistant} & \centering\arraybackslash\small{en} & \centering\arraybackslash\small{*165k}   & \centering\arraybackslash\small{Dialog Act} \\\midrule
\makecell[bl]{\small{ToTTo}\\ \small{\citep{parikh2020totto}}} & \small{Produce an English sentence that describes the highlighted cells in the context of the given table.} & \centering\arraybackslash\small{en} & \centering\arraybackslash\small{136k}  & \centering\arraybackslash\small{Highlighted\quad Table} \\ \midrule
\makecell[bl]{\small{XSum}\\ \small{\citep{narayan2018don}}} & \small{Highlight relevant points in a news article} & \centering\arraybackslash\small{en} & \centering\arraybackslash\small{*25k}  & \centering\arraybackslash\small{Articles} \\\midrule
\makecell[bl]{\small{WebNLG}\\ \small{\citep{gardent2017webnlg}}} & \small{Produce a text that verbalises the input triples in a grammatical and natural way.} & \centering\arraybackslash\small{en/ru} & \centering\arraybackslash\small{50k}   & \centering\arraybackslash\small{RDF triple} \\\midrule
\makecell[bl]{\small{WikiAuto + Turk/ASSET}\\ \small{\citep{jiang-etal-2020-neural}}\\
\small{\citep{alva-manchego-etal-2020-asset}}} & \small{Communicate the same information as the source sentence using simpler words and grammar.}  & \centering\arraybackslash\small{en} & \centering\arraybackslash\small{594k}  & \centering\arraybackslash\small{Sentence} \\\midrule
\makecell[bl]{\small{WikiLingua}\\ \small{\citep{ladhak2020wikilingua}}} & \small{Produce high quality summaries of an instructional article.} & \centering\arraybackslash\small{*ar/cs/de/en es/fr/hi/id/it
ja/ko/nl/pt/ru
th/tr/vi/zh} & \centering\arraybackslash\small{*550k} & \centering\arraybackslash\small{Article}   \\
\bottomrule
\end{tabular}
\caption{A description of all the datasets included in \GEM. The tasks vary in communicative goal, data size, and input type. * indicates changes from the originally published dataset made for \GEM.}
\label{tab:overview}
\end{table*}

This paper describes the selection and construction of the \GEM{} datasets in support of the announcement of the shared task at ACL 2021. More detailed information can be found on our website \url{https://gem-benchmark.com/}.

\section{Benchmarks in NLG}

In this section, we summarize common criticisms of benchmarks in NLP, discuss how they apply to NLG, and how we plan to address them. Then, we describe opportunities that \GEM{} can provide. NLP benchmarks such as GLUE~\citep{wang2018glue} are common for natural language understanding (NLU) tasks. They aggregate multiple tasks under a unified evaluation framework, which enables researchers to fairly compare their models to others. Due to the improved model comparability, benchmarks are critical in measuring modeling progress. 

However, they also pose a risk that progress is reduced to the single number shown in a benchmark's leaderboard and thus may encourage blindly optimizing it without regard to other considerations like model size or fairness~\citep{ethayarajh2020utility}.
This is especially challenging for benchmarks in NLG since, as discussed above, the performance cannot be described through a single metric and it is often not clear what metric to optimize for. This shortfall can be seen in benchmarks like DecaNLP~\citep{mccann2018natural} and GLGE~\citep{liu2020glge} which include NLG tasks but focus only on a single metric and, as a result, may mischaracterize a system's performance.

Moreover, an easy-to-use data infrastructure also disincentivizes researchers from interacting with and conducting in-depth analyses of the data sets that models are trained on. 
The limited analysis delegates the responsibility to ensure that all included datasets have been collected fairly to the creators of the benchmark~\citep{denton2020bringing}. 
The dataset and benchmark creators thus must provide in-depth statements that describe the data characteristics and surface potential issues and consider these issues when selecting datasets for a benchmark~\citep{gebru2018datasheets,bender2018data}.  

These dangers emphasize selecting datasets for a benchmark needs to be carefully done, that the setup has to remain flexible to be able to address newly found limitations, and that the benchmark should focus on climbing a leaderboard. Instead, a living benchmark that can adjust its datasets and specific evaluation metrics can be much more powerful and long-lived. This can, for example, be seen in Dynabench,\footnote{\url{https://dynabench.org/}}~\citep{potts2020dynasent} which has a static evaluation, but interactively adds more test data through a human-in-the-loop approach.

\paragraph{Increasing multilingualism of NLG research.}

Another potentially harmful choice by benchmark creators is the choice of the languages of the included datasets. It is often assumed that work on English transfers to other languages~\citep{Bender2011OnAA}. However, this assumption does not consider differences between the languages that lead to higher modeling complexity, for example, a richer morphology or a flexible word-order. Still, the majority of work in NLP and almost all benchmarks exclusively focus on English~\citep[e.g.,][]{wang2018glue,liu2020glge,mccann2018natural}. Even if multiple languages are considered, the availability of data in a language often does not represent the number of speakers of a language. This means that work on languages with little available data can potentially impact many more people than work on highly resourced languages~\citep{joshi2020state}.

As a result, many recent benchmarking and dataset creation efforts in NLU develop and focus on tasks that are inherently multilingual or which explore cross-lingual transfer. 
For example, XTREME~\citep{hu2020xtreme} introduces a benchmark covering 40 languages across multiple NLU and retrieval tasks, XCOPA~\citep{ponti2020xcopa} is a commonsense reasoning dataset for eleven languages, and MLQA~\citep{lewis2020mlqa} is a dataset for extractive question answering across seven languages. 
We can observe a similar recent trend in natural language generation, where MLSum~\citep{scialom2020mlsum} and WikiLingua~\citep{ladhak2020wikilingua} were created as multilingual summarization datasets. 
There also have been first steps toward including NLG tasks in multilingual NLU benchmarks. For example, XGLUE includes Question and News Title Generation~\citep{liang2020xglue}. Unfortunately, XGLUE reduces the generation evaluation to BLEU-4, a metric that is inadequate for NLG~\citep{reiter2018structured}.

There have also been multiple shared tasks in NLG that focus on multilingualism, for instance, the shared task on multilingual surface realization which includes eleven languages~\citep{mille2018first,mille2019proceedings,mille2020third}. The shared task on document-level generation and translation featured German and English generation challenges~\citep{heafield2020findings}. The WebNLG+ shared task asked participants to contribute models that can realize text in Russian and English~\citep{ferreira20202020}.

A benchmark that focuses only on NLG can enable much richer evaluation (as described in the next sections), and promote non-English datasets. In addition, it can ensure that the datasets created for those shared tasks continue being evaluated. 

\paragraph{Providing a testbed for automated evaluation.} 

Most traditional automated metrics, such as ROUGE~\citep{lin2004rouge} and BLEU~\citep{papineni2002bleu}, measure the n-gram overlap between a reference and the generated text. However, in most cases, there is more than one correct way to generate a text, especially in tasks with a latent content planning or selection step~\citep{reiter2000building}. That means that a correct solution may score low on a metric. While multiple references alleviate the issue somewhat, these metrics still have a low correlation with human judgments~\citep{reiter2018structured, fabbri2020summeval}. 
To address the issue, the machine translation community has been organizing yearly metrics shared tasks which produce metrics that achieve a high correlation~\citep{stanojevic2015results,bojar2016results,bojar2017results,ma2018results,ma2019results,mathur2020results}. The latest metrics focus on semantic equivalence instead of lexical similarity, which improves the correlations drastically. However, recent work by \citet{fabbri2020summeval} demonstrates that this may not hold in summarization, where the automated metric BERTScore~\citep{zhang2019bertscore} does not improve upon the correlation of ROUGE. Moreover, \citet{mathur-2020-tangled} and \citet{freitag-2020-bleu} find that when comparing two high-quality systems, differences according to a metric may also stem from how references are written or flaws in the metric itself.\footnote{For a more complete description of recent developments in NLG evaluation, we refer to the survey by \citet{celikyilmaz2020evaluation}.}

Given that automated metrics perform differently across tasks, setups, and languages, a multi-task NLG benchmark has the opportunity to act as a testbed to evaluate how the latest advances in automated metrics perform on these different tasks. The benchmark can facilitate this research through the release of system outputs and associated human annotations, which is what we are planning to do with \GEM. Moreover, we allow the integration of additional metrics into our living benchmark system, which enables a much faster adoption.

\paragraph{Developing reproducible human evaluation standards.}

In recent work, \citet{howcroft2020twenty} investigated NLG papers from the last twenty years and the evaluation methodologies differ drastically across papers. Moreover, in most cases, it is not even mentioned what the human evaluation aims to measure and that definitions of measures like ``accuracy'' or ``fluency'' are inconsistent. They thus suggest reporting standards for criteria and methods, following a classification system proposed by \citet{belz2020disentangling}. In addition, regularly scheduled shared tasks like WMT have lead to standardization of human evaluation setups and enabled controlled experimentation with them. \GEM{} has the opportunity to develop reproducible standards for how human evaluation for NLG tasks beyond translation should be conducted while at the same time incorporating lessons from related work. Acting on the same need, the recently proposed GENIE~\citep{khasabi2021genie} system aims to automate and standardize the human evaluation of different NLG systems, however with the contrasting goal of reducing the evaluating to a leaderboard-like score. To avoid further fragmentation of the field, \GEM{} is developing its own human evaluation approaches, but uses the infrastructure provided by GENIE to run its human evaluation. 

In addition to GENIE, multiple other related efforts exist that work toward the goal of reproducible and robust in-depth human and automatic evaluation for NLG tasks, and which focus on specific modeling- or task-aspects that are different from those in \GEM. Among those are KILT~\citep{petroni2020kilt} which focuses on knowledge-intensive tasks and retrieval-based models, Storium~\citep{akoury-etal-2020-storium} which focuses on open-ended story generation, and BIG bench\footnote{\url{https://github.com/google/BIG-bench}} which focuses on measuring few-shot and zero-shot capabilities of language models. 

\section{Dataset Selection}
\label{sec:selection}

As highlighted in Figure~\ref{fig:circle}, the selection of included datasets is an integral part of a benchmark. They should be challenging for models, but it should still be possible to evaluate models trained on them. Moreover, the datasets should cover a wide range of relevant generation challenges that allow for findings to be as general as possible. Finally, the datasets should cover tasks that are interesting for contributors to work on to facilitate the wide adoption of the benchmark. 

To collect datasets with those desired properties, the selection methodology for \GEM{} is composed of three steps. First, we elicited a set of proposals from everyone involved in the effort. Second, we identified criteria for the selection. Third, all \GEM{} members voted on individual dataset and criteria utilities. The final selection maximizes the utility under constrained resources, similar to a knapsack solver.\footnote{Consider the criterion ``We need equal representation of large and small datasets'' under the constraint that only two datasets can be selected. If we have two large datasets with utility 10, and one small one with utility 5, we may want to include the smaller dataset over the second large dataset to satisfy the criterion.}
This can be seen as an extension of the selection process of SuperGLUE~\citep{wang2019superglue} that had similar first and second steps but made the final decision based on which were harder for a baseline model to solve after identifying a final set of candidate datasets. Since we are going to introduce challenge sets, the baseline performance of models on a dataset matters less.

\paragraph{Dataset Elicitation.} In the first step, all \GEM{} participants were asked to suggest datasets following the schema provided in Appendix~\ref{app:suggestion}. The categories included multiple brief categorizations, such as a description of the challenge that this dataset provides, its high-level task, and the communicative goal of an agent trained on the data. Following our goal to focus on non-English languages, we further asked for the languages included in the dataset, as well as the language locale. This step yielded 35 proposed datasets, listed in Appendix~\ref{app:considered}.

\paragraph{Estimating Task+Criterion Utility.} The second step focused on the selection of criteria to inform the selection. The initial set of criteria was selected through open discussion involving all members.
We split criteria into ``hard'' and ``soft'' ones -- hard criteria would lead to the definite inclusion/exclusion of a task if (not) satisfied. Soft criteria inform the utility of the remaining tasks. All \GEM{} members filled out a survey asking them to rate, on a 5-point Likert scale, how much they wanted to see a task included in \GEM. Additionally, we posed yes/no questions for all considered hard criteria and various questions about the soft criteria (e.g., ``what percentage of the tasks should feature non-English language?'', or ``do we prefer noisy or clean datasets?''). 
Finally, the survey included open text fields that asked for (1) comments on any of the tasks, (2) comments or suggestions on hard exclusion criteria, and (3) suggestions of additional criterion/criteria. The full list of questions is shown in Appendix~\ref{app:survey}.

The survey received 28 responses, revealing that the initial version of GEM should include a median of 10 tasks or an average of 12. Of those tasks, about a third should feature non-English language. 

\paragraph{Selected Criteria.} For the hard criteria, there was an agreement to focus only on open-access datasets and that concurrent or past shared tasks for the same datasets are not an issue. Overall, the sentiment determined the following selection principles: 
\begin{itemize}[itemsep=0mm]
    \item We focus on diverse high-level tasks over a single high-level task evaluated in-depth. However, each high-level task should include multiple datasets. 
    \item We focus on clean datasets to avoid conflating model mistakes and learned noise.
    \item We include a mix of high- and low-resource datasets.
    \item We focus on data with interesting test sets. 
    \item We should not focus on the quality of current evaluation strategies for a given dataset.
    \item We prefer multi-reference datasets since those have been shown to lead to more robust automatic evaluation.
\end{itemize}

\paragraph{High-Level Tasks.} Since these principles dictate that we should focus on a small set of high-level tasks, we used the free-text replies to evaluate the interest in different high-level tasks. Grouping the proposed tasks yielded the following candidates: Summarization, Dialog, Simplification/Compression, Question Answering, Creative Writing, Data-to-Text, and Question Generation.\footnote{For a full overview of potential future expansions and challenges, we refer to the survey by \citet{gatt2018survey}.}  
There was a preference to exclude image inputs and question answering because those tasks add complexity to the evaluation beyond the generated text. 
Moreover, since creative generation tasks like story generation and poetry generation suffer even more from inadequate evaluation approaches, there was a consensus to not include them.
There was, however, a strong preference for the high-level tasks Summarization, Data-to-text, and Dialog.\footnote{One may question the absence of Translation from this list. While it is a generation task, we excluded it since Translation already has regular benchmarking efforts with WMT.} 

\paragraph{Specific Datasets.} The final selection is shown in Table~\ref{tab:overview}. To arrive at the selection, we first ranked all datasets by their average rating. For this, we treated positive ratings as 1, negative ratings as -1, and neutral ratings as 0. The highest-ranked datasets were E2E with 0.577, XSum with 0.538, and ToTTo with 0.461. Unfortunately, non-English datasets were ranked lower, with only WebNLG and MLSum among the top 15 datasets. 
We grouped all datasets by their high-level tasks and selected a group that would not violate the selection principles (e.g., only high-resource tasks). If two datasets fit, we picked the one with a higher interest rating. 
Among the 11 datasets, we have 18different languages, and the dataset sizes range from 5,000 examples to 1.5M, with most datasets between 50-150k examples. Two of them do not include English at all, which we hope reduces the dependence of the modeling approaches on anglocentric pretraining~\citep{anastasopoulos2020cross}. The high-level tasks include Dialog, Summarization, Data-to-Text, and Simplification. About half of the datasets have multiple references and more than half had post-processing steps applied to them to ensure high data quality.

\subsection{GEMifying the data}
We produce data cards~\citep{bender2018data,gebru2018datasheets} for all data sets in \GEM, for which we developed an NLG-specific template.\footnote{Our template extends and restructures that from Hugging Face Datasets and along with a guide can be found at \url{https://gem-benchmark.com/data_cards}.} In addition to describing the data itself, the cards acknowledge potential limitations of a dataset regarding its creation process and describe its real-world use cases to ensure that the research is conducted responsibly.

These datasets are the base selection, and as part of \GEM, we may change datasets and how they are used. For example, we may improve the training sets, make the test sets more challenging, or probe for specific skills a model must exhibit with test-only datasets~\citep{perez2017analysing,linzen2020accelerate,ribeiro-etal-2020-beyond,schlegel2020beyond}. 
We may also ask to evaluate a single model on multiple test sets, following the design by \citet{dua2019orb}.

\begin{table*}[t]
\begin{tabular}{@{}llm{15em}@{}}
\toprule
Challenge Set Type & Example & Tasks \\ \midrule
Numerical Variation & 53 -\textgreater 79 & WebNLG \\
Attribute Order & English Cheap -\textgreater Cheap English & All data-to-text tasks \\
Typographical Errors & English Cheap -\textgreater Enlish Chesp & Schema-Guided, WikiAuto, XSum \\
No Punctuation & ... the dog. -\textgreater ... the dog & Schema-Guided, WikiAuto, XSum \\
Backtranslation & fantastic -\textgreater toll -\textgreater great & Schema-Guided, WikiAuto, XSum \\ \midrule
Train \& Validation Samples &  & All tasks \\
Gender, Ethnicity, Nationality &  & ToTTo \\
Input Shape &  & WebNLG \\
Syntactic Complexity &  & WikiAuto \\ \midrule
Covid Summaries &  & MLSUM (es+de), XSum \\ \bottomrule 
\end{tabular}
\caption{An overview of the types of challenge sets for GEM. The first category are modifications to inputs of a model, the second category identifies contrast sets which are subsets of the original test set, and the third describes newly collected data.}
\label{tab:challenge-sets}
\end{table*}

We are including modifications to several of the datasets: (1) \textbf{MLSum}: We excluded all languages besides Spanish and German since the sources for other languages disallow scraping content. Additionally, we removed all duplicate items (i.e., items with the same input text) and we used langdetect\footnote{\url{https://pypi.org/project/langdetect/}} to filter out examples that were in the wrong language. In total, 147 examples were removed from the German portion (0.06\%) and 7417 examples were removed from the Spanish portion (2.5\%).
(2) \textbf{XSum}: Summaries in this dataset often have divergence issues between the source and target texts since gold summaries are introductory sentences prefacing each article. Models agnostic to such noises are vulnerable to hallucinations~\citep{wiseman-etal-2017-challenges,dhingra2019handling}. To combat this, we fine-tuned a BERT-based~\citep{devlin-2019-bert} classifier on 500 document and gold summary pairs, manually annotated for faithfulness~\citep{maynez2020faithfulness} and excluded all document-summary pairs from the original XSum dataset where the classifier was not confident ($p(\textrm{faithful}) > 0.8$) whether the summary is faithful to the document or not.
(3) \textbf{Schema-Guided Dialog}: We are focusing on the response-generation part of the dataset and thus reformatted the dataset to treat the service agent utterances as the targets to be generated and the previous customer utterance and the agent's dialog act as the input. We additionally reformat the dialog acts to directly conform to the format described in the paper~\citep{kale2020few}. 
(4) \textbf{WikiLingua}: We focus on the same five languages that were benchmarked in its original release (en, es, ru, tr, vi) in a cross-lingual setup in which the inputs are in the respective language and the outputs are in English. However, we re-split the original data to avoid train-test overlaps between languages and provide training data in 13 additional languages (as shown in Table~\ref{tab:overview}). For \GEM, we allow submissions trained on any of the languages in isolation or as part of a multilingual model.

\subsection{Challenge Sets} 

In addition to applying consistent metrics to existing test sets, understanding specific model behavior, such as model generalization capabilities or performance under targeted cases, is also key for improvement. This is difficult to assess through evaluations on i.i.d. test splits. We thus release challenge sets to evaluate data-to-text and text-to-text models (overview in Table~\ref{tab:challenge-sets}). In addition to enabling a more specific breakdown of how a model performs in the presence of challenging inputs, the set of system outputs on these test sets also constitutes a rich corpus that enables further error analysis and research.
We apply multiple strategies to create the special test sets, in particular (I) alteration of the existing test sets (e.g., the introduction of distractors), (II) breaking down of the existing sets into subsets with certain properties (e.g., subsets with different complexity), and (III) the compilation of new test sets (e.g., out-of-vocabulary inputs).
We restrict the size of each challenge set to about 500 examples to minimize computational overhead. On the WebNLG challenge sets, all subset items are selected proportionally from each category to ensure a similar distribution to the original set; on all other datasets the subset items are selected from the whole set. The results of the different systems on these subsets will be compared to the results obtained by the same systems on the same subsets of the original test data. 

For case (I), altering existing test sets, the first challenge set adds \textbf{numerical variation} in WebNLG. This variation attempts to respect the format of the current cardinal value (e.g. alpha, integer, or floating-point) and replaces the existing value with a new random value as a means to challenge existing trained models. The generated number is lower-bounded between zero and upper bounded to be within to the highest power of 10 unit for the given value (e.g. replacing a value of 54 would result in a random value between 0-100). Floating values are also bounded to have the same degree of precision as the input value.  
For structure-to-text and dialog datasets, we produce a version of the test sets in which the \textbf{order of the components} of the input structures (triples, concepts, dialog acts, table rows, etc.) is randomly changed.
For text-to-text datasets and Schema-guided Dialog, we introduce several types of perturbations: (a) \textbf{typographical errors}, using butter-fingers \footnote{\url{https://github.com/alexyorke/butter-fingers}} with two thresholds $0.02$ and $0.05$, which respectively correspond to lower and higher error frequencies; (b) \textbf{removal of the final punctuation} sign (if any); (c) substitution of the input text by a \textbf{backtranslated version}, using the backtranslation implementation by \citet{xie2020unsupervised}. We rejected backtranslation outputs based on a character length to ensure that the difference in character length between original and backtranslation does not exceed 35\% of the original source character length. For XSum 99.8\% of the backtranslations were accepted, for WikiAuto 94.42\% (ASSET) and 87.18\% (TURK), and for Schema-Guided Dialog 78\%.

In case (II), the breaking down existing sets, we first provide for each dataset \textbf{random samples of training and validation data}, in order to assess to what extent the scores of the different systems drop when run on the test data. Then, specific splits are created for particular datasets, in order to assess possible biases of the models, and their robustness across inputs with different specifications. For ToTTo, test set splits are built according to several aspects that can be identified using WikiData: \textbf{gender, ethnicity and nationality grouped by continent}. For gender, we compare the performance between male and female people, but cannot compare other genders due to a lack of original data - only seven people in the original test set are marked as having a different gender. We compare across the continent of the underlying nationality to address the issue that data for each country can be very sparse -- i.e., only 19 countries are represented by more than ten people and only one of these is located in Africa (Kenya). In case a person has citizenships across multiple continents, we may include the person in any of the included continents. Finally, we compare African Americans vs. all Americans. Ethnicity is very sparsely annotated in WikiData with fewer than 150 annotated test examples  in total and 128 of these are African Americans. We thus are unable to compare the performance on, e.g., Yoruba or Punjabi people, both of which have fewer than five instances. Another caveat here is that only 21 of the 128 people are female. Our contrast subset that can include any US citizens matches these counts. Across all three challenge subsets, we additionally match the fraction of the existing non-overlap and overlap properties. 
For WebNLG, we propose subsets based on the \textbf{shape of the inputs} (number of triples, number of common subjects and/or objects, depth, etc.) For Turk/ASSET, splits are created in terms of the \textbf{syntactic complexity} of the sentences to be simplified. To characterise sentence complexity we use the developmental level scale proposed by \citet{covington2006complex}.\footnote{We use the implementation provided by \citet{lu2010automatic}.} 
For all datasets, we propose splits based on the frequency of the parts that compose the input in the training data; the resulting test sets range from being made of very common components to being made only from components unseen in the training data.
For case (III), we collect time-shifted test data for news summarization in the form of articles with Covid19-related keywords. Since MLSum and XSum were collected before the pandemic, we can measure how a model responds to context not seen in the training data (outside of potential pretraining). The new set of articles covers existing article topics (economy, sports, etc.) but all in relation to the Covid19 pandemic. In addition, some new topics appear in the collected data derived from outlet sections that were not part of the original data collection.\footnote{To collect this data we use the scripts provided for the re-creation of MLSum and XSum datasets.}

\section{Experimental Setup}

Since the \GEM{} test sets and final metrics selection have not been released yet, we describe an experimental setup that will ensure that participating models are trained correctly and evaluated on publicly available data with available metrics that will give a sufficient indication of a model's performance. To do this, we are reporting the results of the baseline models on the validation sets.

\subsection{Modeling Baselines}

Much of the recent modeling progress in NLP can be attributed to the rise of the pretrain-then-finetune paradigm which has led to consistently better results. This finding is consistent with human judgments for summarization, as shown by \citet{fabbri2020summeval}, among others. 
However, many of the tasks included in \GEM{} may not benefit from a language model encoder since their input is not natural language. We thus apply a variety of different architectures that vary in size, complexity, and training schema. 
Our main baselines are T5 with 60M parameters~\citep{raffel2019exploring} and BART with 139M parameters~\citep{lewis2019bart}. For non-English datasets, we use their multilingual counterparts mT5 in various sizes~\citep{xue2020mt5} and mBART~\citep{liu2020multilingual}. 
We additionally train the following baselines on a subset of tasks: TGen (with added language model and lemma tags denoted as TGen+/++)~\citep{dusek2016sequence}, an architecture for generation from dialog acts, an LSTM-based Sequence-to-sequence model with attention~\citep{bahdanau2014neural}, DialoGPT~\citep{zhang2019dialogpt}, a pretraining approach for conversational models, and PEGASUS~\citep{zhang2020pegasus}, which uses a summarization-specific pretraining schema that masks and predicts entire sentences.
For WikiLingua, we additionally report results on a setup proposed by \citet{ladhak2020wikilingua} which includes first training a monolingual model followed by finetuning with the correct source language, coupled with synthetic data generated through translation (mBART+). 

\noindent Almost all baselines can be reproduced on a GPU-based colaboratory notebook within 2-3 hours. 

\subsection{Automated Evaluation}

As mentioned above, \GEM{} provides a testbed for automated metrics and can be used to popularize newly developed ones. Thus, models are evaluated via a constantly expanding list of metrics and, to avoid overfitting to known metrics, we will use metrics on the test submissions that are not included in this initial writeup. Consequentially, the baseline results are an incomplete list which will be expanded upon the announcement of the test metrics. 
The set of metrics can be computed via the framework described at \url{https://gem-benchmark.com/shared_task} which comprises metrics in the following categories:

\begin{figure*}[t]
\centering
\includegraphics[width=\linewidth]{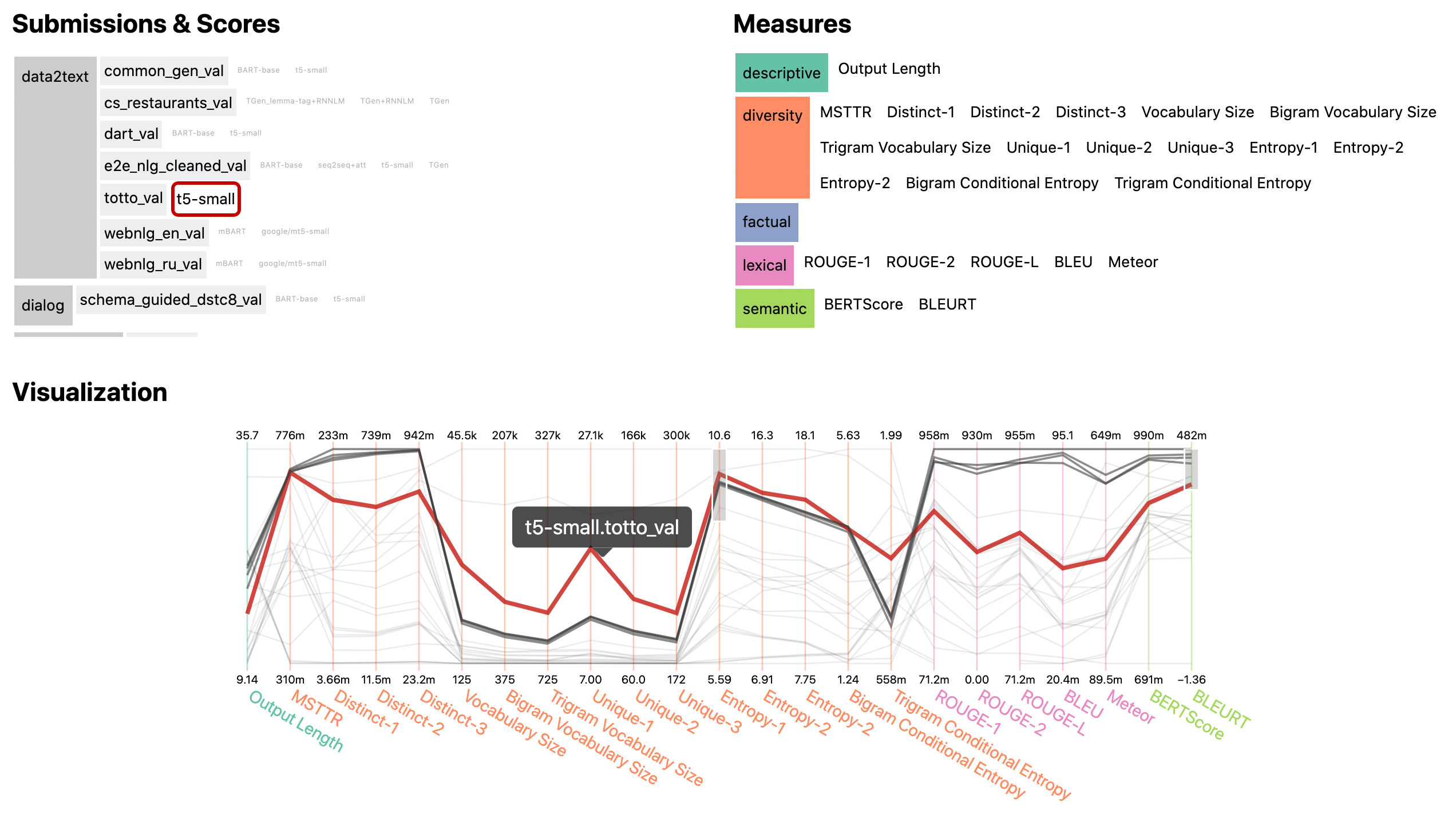}
\vspace{-1cm}
\caption[Interactive result exploration]{A screenshot of the interactive result exploration tool. \textbf{[Top Left]} The selection of tasks, task-groups, or individual submissions. \textbf{[Top Right]} The selection of metric-groups or metrics \textbf{[Bottom]} The parallel coordinates visualization of the selection. The selection here can be filtered by brushing over a section of an individual metric, as is shown here for BLEURT. Hovering over a line presents detailed information of the particular submission.}
\label{fig:results}
\end{figure*}

\paragraph{Lexical Similarity.} 
We include multiple ``traditional'' metrics as baseline metrics, notably BLEU~\citep{papineni2002bleu}, ROUGE-1/2/L~\citep{lin2004rouge}, and METEOR~\citep{banerjee2005meteor}. 
These metrics can often be gamed, for example, ROUGE can be improved by increased the output length of the model~\citep{sun2019compare}. Moreover, the reliability of these metrics depends on the quality and number of the references~\citep{mathur-2020-tangled,freitag-2020-bleu}. However, on a system-level, they still correlate well with human judgments for some tasks~\citep{reiter2018structured}.

\paragraph{Semantic Equivalence.} More recently, metrics that rely on pretrained language models have shown improved correlations with human judgments on the segment-level. We thus include BERTScore~\citep{zhang2019bertscore}, a metric based on the similarity of sentence embeddings, and BLEURT~\citep{sellam-2020-bleurt}, a metric that is fine-tuned on human ratings. The reported baseline results use RoBERTa-large~\citep{liu2019roberta} and mBERT~\citep{devlin-2019-bert} for BERTScore and the English-only BLEURT-base-128 for BLEURT.

\renewcommand{\arraystretch}{0.75}
\begin{table*}[!htbp]
\centering
\small
\begin{tabular}{@{}llrrrrrrr@{}}
\toprule
\multirow{2}{*}{\textbf{Dataset}} & \multirow{2}{*}{\textbf{Model}} & \multicolumn{7}{c}{\textbf{Metrics {\fontsize{9}{10}\selectfont (Lexical Similarity and Semantic Equivalence)}}} \\ 
 &  & {\fontsize{7.5}{9}\selectfont \textbf{METEOR}} & {\fontsize{7.5}{9}\selectfont \textbf{ROUGE-1}} & {\fontsize{7.5}{9}\selectfont \textbf{ROUGE-2}} & {\fontsize{7.5}{9}\selectfont \textbf{ROUGE-L}} & {\fontsize{7.5}{9}\selectfont \textbf{BLEU}} & {\fontsize{7.5}{9}\selectfont \textbf{BERTScore}} & {\fontsize{7.5}{9}\selectfont \textbf{BLEURT}} \\ \toprule
\multirow{2}{*}{CommonGen} & \small{BART} & 0.301 & 63.5 & 32.5 & 55.1 & 27.5 & 0.943 & -0.400 \\
 & \small{T5} & 0.291 & 64.0 & 29.4 & 54.5 & 26.4 & 0.942 & -0.412 \\
\midrule
\multirow{3}{*}{Czech Restaurant} & \small{mT5-small} & 0.229 & 47.3 & 28.6 & 43.0 & 17.9 & 0.895 & -- \\
& \small{mT5-base} & 0.23 & 48.1 & 28.8 & 44.2 & 17.1 & 0.898 & -- \\
& \small{mT5-large} & 0.233 & 51.3 & 30.0 & 46.4 & 17.5 & 0.902 & -- \\
& \small{mT5-XL} & 0.229 & 52.1 & 31.3 & 47.3 & 17.0 & 0.905 & -- \\
 & \small{TGen} & 0.152 & 13.6 & 0.0 & 13.6 & 0.03 & 0.650 & -- \\
 & \small{TGen+}  & 0.151 & 13.8 & 0.0 & 13.8 & 0.03 & 0.651 & --  \\
 & \small{TGen++} & 0.167 & 9.7 & 0.0 & 9.7 & 0.03 & 0.648 & -- \\
\midrule
\multirow{2}{*}{DART} & \small{BART} & 0.107 & 7.1 & 0.0 & 7.1 & 0.02 & 0.862 & -0.261 \\
 & \small{T5} & 0.115 & 8.4 & 0.0 & 8.4 & 0.02 & 0.901 & -0.091 \\
\midrule
\multirow{4}{*}{E2E clean} & \small{BART} & 0.373 & 73.6 & 48.5 & 57.8 & 43.5 & 0.948 & 0.190 \\
 & \small{LSTM} & 0.394 & 75.0 & 50.3 & 58.9 & 46.9 & 0.950 & 0.252 \\
 & \small{T5} & 0.369 & 72.6 & 47.5 & 56.4 & 43.0 & 0.945 & 0.384 \\
 & \small{TGen} & 0.391 & 74.7 & 49.6 & 58.4 & 46.0 & 0.949 & 0.412 \\
\midrule
\small{MLSum (de)} & \small{mBART} & 0.437 & 43.8 & 33.1 & 39.8 & 28.2 & 0.888 & -- \\
& \small{mT5-small} & 0.098 & 11.8 & 3.4 & 10.0 & 5.0 & 0.826 & -- \\

& \small{mT5-base} & 0.099 & 12.2 & 3.5 & 10.2 & 5.1 & 0.830 & -- \\
& \small{mT5-large} & 0.101 & 12.4 & 3.6 & 10.4 & 5.2 & 0.832 & -- \\
& \small{mT5-XL} & 0.102 & 12.6 & 3.7 & 10.5 & 5.3 & 0.832 & -- \\
\small{MLSum (es)} & \small{mBART} & 0.210 & 28.4 & 10.9 & 22.4 & 7.4 & 0.836 & -- \\
& \small{mT5-small} & 0.198 & 28.1 & 10.5 & 22.8 & 8.2 & 0.834 & -- \\
& \small{mT5-base} & 0.214 & 29.5 & 11.7 & 23.9 & 9.6 & 0.839 & -- \\
& \small{mT5-large} & 0.235 & 31.8 & 13.8 & 26.0 & 11.0 & 0.845 & -- \\
& \small{mT5-XL} & 0.247 & 33.1 & 15.0 & 27.2 & 11.9 & 0.849 & -- \\
\midrule
\multirow{2}{*}{Schema-Guided} & \small{BART} & 0.089 & 13.6 & 4.4 & 11.3 & 2.7 & 0.691 & -1.355 \\
 & \small{T5} & 0.331 & 58.2 & 36.8 & 52.6 & 33.4 & 0.874 & 0.009  \\
\midrule
\small{ToTTo} & \small{T5} & 0.363 & 70.1 & 48.3 & 60.1 & 42.2 & 0.914 & 0.179 \\
\midrule
\small{XSum} & \small{PEGASUS} & 0.216 & 46.5 & 23.2 & 38.1 & 17.0 & 0.918 & -0.186 \\
\midrule
\multirow{2}{*}{WebNLG (en)} & \small{mBART} & 0.462 & 83.4 & 63.1 & 70.3 & 66.1 & 0.967 & 0.458 \\
& \small{mT5-small} & 0.442 & 78.8 & 59.2 & 67.2 & 60.2 & 0.948 & 0.416  \\
& \small{mT5-base} & 0.461 & 82.3 & 62.1 & 69.7 & 65.2 & 0.955 & 0.451 \\
& \small{mT5-large} & 0.473 & 83.8 & 64.4 & 71.6 & 68.0 & 0.959 & 0.479 \\
& \small{mT5-XL} & 0.472 & 83.5 & 63.6 & 71.0 & 67.6 & 0.958 & 0.47 \\
\multirow{2}{*}{WebNLG (ru)} & \small{mBART} & 0.613 & 34.8 & 13.4 & 33.0 & 47.0 & 0.888 & -- \\
& \small{mT5-small} & 0.553 & 29.7 & 10.5 & 28.4 & 41.1 & 0.942 & -- \\
& \small{mT5-base} & 0.602 & 33.0 & 12.7 & 31.3 & 44.3 & 0.949 & -- \\
& \small{mT5-large} & 0.614 & 33.4 & 13.4 & 32.1 & 46.4 & 0.952 & -- \\
& \small{mT5-XL} & 0.624 & 34.3 & 13.7 & 32.8 & 47.2 & 0.952 & -- \\
\midrule
\multirow{2}{*}{Turk} & \small{BART} & 0.556 & 90.3 & 86.1 & 89.9 & 88.3 & 0.967 & 0.358 \\
 & \small{T5} & 0.649 & 95.7 & 92.9 & 95.5 & 95.1 & 0.974 & 0.495 \\
\midrule
\multirow{2}{*}{ASSET} & \small{BART} & 0.560 & 90.1 & 82.3 & 89.6 & 92.4 & 0.982 & 0.407 \\
 & \small{T5} & 0.581 & 92.1 & 92.3 & 92.6 & 93.4 & 0.984 & 0.468 \\
\midrule
\multirow{2}{*}{WikiLingua (es$\rightarrow$en)} & \small{mBART} & 0.178 & 38.3 & 15.4 & 32.4 & 12.2 & 0.853 & -0.290 \\
 & \small{mBART+} & 0.196 & 40.7 & 16.9 & 34.1 & 14.3 & 0.858 & -0.248 \\
 & \small{mT5-small} & 0.135 & 29.8 & 9.8 & 25.5 & 7.4 & 0.832 & -0.437 \\
& \small{mT5-base} & 0.162 & 36.3 & 13.7 & 30.6 & 10.1 & 0.85 & -0.324 \\
& \small{mT5-large} & 0.183 & 39.3 & 15.7 & 33.0 & 12.5 & 0.857 & -0.27 \\
& \small{mT5-XL} & 0.203 & 41.8 & 17.4 & 34.7 & 15.2 & 0.862 & -0.218 \\

\multirow{2}{*}{WikiLingua (ru$\rightarrow$en)} & \small{mBART} & 0.153 & 33.1 & 11.9 & 27.8 & 9.3 & 0.839 & -0.369 \\
 & \small{mBART+} & 0.174 & 37.3 & 14.9 & 31.9 & 12.0 & 0.851 & -0.303 \\
 & \small{mT5-small} & 0.128 & 27.2 & 8.5 & 23.2 & 6.9 & 0.825 & -0.471 \\
& \small{mT5-base} & 0.149 & 32.5 & 11.1 & 26.9 & 8.8 & 0.839 & -0.377 \\
& \small{mT5-large} & 0.167 & 35.0 & 12.7 & 28.8 & 11.0 & 0.846 & -0.337 \\
& \small{mT5-XL} & 0.185 & 38.6 & 15.4 & 32.3 & 13.6 & 0.855 & -0.268 \\
\multirow{2}{*}{WikiLingua (tr$\rightarrow$en)} & \small{mBART} & 0.164 & 34.4 & 13.0 & 28.1 & 11.7 & 0.837 & -0.414 \\
 & \small{mBART+} & 0.204 & 43.7 & 20.8 & 37.9 & 17.5 & 0.866 & -0.252 \\
 & \small{mT5-small} & 0.154 & 29.4 & 10.9 & 23.4 & 13.0 & 0.823 & -0.595 \\
& \small{mT5-base} & 0.168 & 32.5 & 13.6 & 26.0 & 15.5 & 0.834 & -0.507 \\
& \small{mT5-large} & 0.185 & 36.2 & 15.0 & 29.1 & 16.9 & 0.846 & -0.405 \\
& \small{mT5-XL} & 0.208 & 41.5 & 19.6 & 34.7 & 19.9 & 0.86 & -0.291 \\
\multirow{2}{*}{WikiLingua (vi$\rightarrow$en) } & \small{mBART} & 0.150 & 32.0 & 11.1 & 26.4 & 9.2 & 0.836 & -0.394 \\
 & \small{mBART+} & 0.183 & 38.1 & 15.4 & 32.5 & 13.3 & 0.853 & -0.284 \\
 & \small{mT5-small} & 0.12 & 23.5 & 6.0 & 19.0 & 6.1 & 0.812 & -0.56 \\
& \small{mT5-base} & 0.129 & 26.0 & 7.5 & 20.5 & 7.4 & 0.82 & -0.513 \\
& \small{mT5-large} & 0.146 & 29.9 & 9.6 & 23.8 & 9.2 & 0.833 & -0.421 \\
& \small{mT5-XL} & 0.173 & 35.5 & 13.0 & 29.2 & 12.4 & 0.847 & -0.308 \\
\bottomrule
\end{tabular}
\caption{The set of baseline results we release alongside \GEM{} with a focus on reference-based evaluation.}
\label{tab:results}
\end{table*}

\setlength{\tabcolsep}{4pt}
\renewcommand{\arraystretch}{0.75}
\begin{table*}[!htbp]
\centering
\small
\begin{tabular}{@{}llrrrrrrrrr@{}}
\toprule
\multirow{2}{*}{\textbf{Dataset}} & \multirow{2}{*}{\textbf{Model}} & \multicolumn{9}{c}{\textbf{Metrics {\fontsize{9}{10}\selectfont (Diversity and System Characterization)}}} \\ 
 &  & {\fontsize{7.5}{9}\selectfont \textbf{MSTTR}} & {\fontsize{7.5}{9}\selectfont \textbf{Distinct$_1$}} & {\fontsize{7.5}{9}\selectfont \textbf{Distinct$_2$}} & {\fontsize{7.5}{9}\selectfont \textbf{$H_1$}} & {\fontsize{7.5}{9}\selectfont \textbf{$H_2$}} & {\fontsize{7.5}{9}\selectfont \textbf{Unique$_1$}} & {\fontsize{7.5}{9}\selectfont \textbf{Unique$_2$}}& {\fontsize{7.5}{9}\selectfont \textbf{$|\mathcal{V}|$}} & {\fontsize{7.5}{9}\selectfont \textbf{Output Len.}} \\ \toprule
\multirow{2}{*}{CommonGen} & \small{BART} & 0.57 & 0.12 & 0.41 & 7.1 & 10.7 & 583 & 2.7k & 1.2k & 10.5 \\
 & \small{T5} & 0.51 & 0.11 & 0.36 & 6.5 & 10.1 & 465 & 2.0k & 1.0k & 9.6 \\
\midrule
\multirow{3}{*}{Czech Restaurant} & \small{mT5-small} & 0.51 & 0.04 & 0.1 & 6.2 & 7.8 & 86 & 278 & 287 & 10.2 \\
 & \small{mT5-base} & 0.49 & 0.03 & 0.09 & 6.1 & 7.6 & 80 & 249 & 273 & 10.5 \\
 & \small{mT5-large} & 0.57 & 0.05 & 0.13 & 6.6 & 8.4 & 103 & 387 & 361 & 10.1 \\
 & \small{mT5-XL} & 0.6 & 0.06 & 0.19 & 6.8 & 9.0 & 146 & 614 & 438 & 9.5 \\
 & \small{TGen} & 0.57 & 0.03 & 0.11 & 6.4 & 8.0 & 58 & 239 & 245 & 9.1 \\
 & \small{TGen+} & 0.61 & 0.04 & 0.12 & 6.5 & 8.1 & 84 & 290 & 305 & 9.2 \\
 & \small{TGen++} & 0.56 & 0.04 & 0.11 & 6.5 & 8.1 & 85 & 280 & 297 & 9.5 \\
\midrule
\multirow{2}{*}{DART} & \small{BART} & 0.55 & 0.19 & 0.45 & 8.4 & 11.3 & 1.3k & 3.6k & 2.4k & 12.0 \\
 & \small{T5} & 0.51 & 0.19 & 0.42 & 8.0 & 10.7 & 1.2k & 3.1k & 2.1k & 10.8 \\
\midrule
\multirow{4}{*}{E2E clean} & \small{BART} & 0.32 & 0.005 & 0.02 & 5.7 & 7.2 & 16 & 104 & 149 & 22.0 \\
 & \small{LSTM} & 0.31 & 0.004 & 0.02 & 5.6 & 7.1 & 19 & 106 & 139 & 23.1 \\
 & \small{T5} & 0.30 & 0.004 & 0.01 & 5.6 & 6.9 & 7 & 60 & 125 & 23.0 \\
 & \small{TGen} & 0.31 & 0.004 & 0.02 & 5.6 & 7.2 & 19 & 116 & 140 & 23.2 \\
\midrule
\small{MLSum (de)} & \small{mBART} & 0.78 & 0.11 & 0.52 & 10.6 & 16.3 & 27k & 166k & 46k & 35.7 \\
 & \small{mT5-small} & 0.75 & 0.12 & 0.52 & 10.4 & 15.8 & 20.1k & 113.8k & 33.6k & 24.7 \\
 & \small{mT5-base} & 0.76 & 0.12 & 0.53 & 10.4 & 15.8 & 20.2k & 113.0k & 33.3k & 24.2 \\
 & \small{mT5-large} & 0.76 & 0.12 & 0.53 & 10.4 & 15.8 & 20.0k & 114.0k & 33.3k & 24.4 \\
 & \small{mT5-XL} & 0.77 & 0.12 & 0.53 & 10.4 & 15.8 & 20.0k & 114.6k & 33.3k & 24.5 \\
\small{MLSum (es)} & \small{mBART} & 0.71 & 0.10 & 0.47 & 10.1 & 15.7 & 19k & 120k & 35k & 32.3 \\
 & \small{mT5-small} & 0.69 & 0.12 & 0.48 & 10.0 & 15.1 & 14.0k & 77.6k & 25.5k & 21.7 \\
 & \small{mT5-base} & 0.71 & 0.12 & 0.5 & 10.1 & 15.3 & 15.1k & 85.2k & 27.2k & 23.0 \\
 & \small{mT5-large} & 0.71 & 0.12 & 0.5 & 10.1 & 15.3 & 14.9k & 82.0k & 26.6k & 22.1 \\
 & \small{mT5-XL} & 0.72 & 0.12 & 0.5 & 10.1 & 15.3 & 14.8k & 80.5k & 26.1k & 21.4 \\
\midrule
\multirow{2}{*}{Schema-Guided} & \small{BART} & 0.56 & 0.02 & 0.06 & 7.0 & 9.2 & 1.8k & 6.2k & 3.9k & 22.0 \\
 & \small{T5}  & 0.67 & 0.03 & 0.10 & 7.9 & 10.6 & 1.6k & 5.8k & 3.8k & 11.8 \\
\midrule
\small{ToTTo} & \small{T5} & 0.73 & 0.18 & 0.54 & 10.1 & 14.4 & 15k & 60k & 21k & 15.3 \\
\midrule
\small{XSum} & \small{PEGASUS} & 0.73 & 0.20 & 0.64 & 9.3 & 13.1 & 3.0k & 13k & 5k & 22.9 \\
\midrule
\small{WebNLG (en)} & \small{mBART} & 0.53 & 0.09 & 0.27 & 8.6 & 11.8 & 969 & 4.0k & 3.2k & 20.7 \\
 & \small{mT5-small} & 0.5 & 0.09 & 0.25 & 8.6 & 11.8 & 864 & 3.9k & 3.2k & 22.7 \\
 & \small{mT5-base} & 0.53 & 0.09 & 0.27 & 8.7 & 11.9 & 983 & 4.4k & 3.3k & 21.7 \\
 & \small{mT5-large} & 0.54 & 0.09 & 0.29 & 8.7 & 12.0 & 1.1k & 4.8k & 3.4k & 21.7 \\
 & \small{mT5-XL} & 0.54 & 0.09 & 0.29 & 8.7 & 12.0 & 1.1k & 4.8k & 3.4k & 21.6 \\
\small{WebNLG (ru)} & \small{mBART} & 0.46 & 0.08 & 0.20 & 8.1 & 10.3 & 334 & 1.1k & 1.2k & 18.9 \\
 & \small{mT5-small} & 0.43 & 0.08 & 0.20 & 7.9 & 10.2 & 349 & 1.2k & 1.2k & 19.2 \\
 & \small{mT5-base} & 0.47 & 0.09 & 0.23 & 8.2 & 10.7 & 482 & 1.6k & 1.4k & 19.9 \\
 & \small{mT5-large} & 0.48 & 0.09 & 0.24 & 8.2 & 10.7 & 474 & 1.6k & 1.4k & 19.4 \\
 & \small{mT5-XL} & 0.46 & 0.09 & 0.22 & 8.2 & 10.5 & 418 & 1.4k & 1.3k & 19.5 \\
 \midrule
\multirow{2}{*}{Turk} & \small{BART} & 0.73 & 0.23 & 0.74 & 9.8 & 14.1 & 5.5k & 23k & 8.6k & 18.4 \\
 & \small{T5} & 0.73 & 0.22 & 0.72 & 9.9 & 14.2 & 5.9k & 25k  & 9.3k & 20.1 \\
\midrule
\multirow{2}{*}{ASSET} & \small{BART} & 0.73 & 0.23 & 0.73 & 9.8 & 14.1 & 5.9k & 24k & 9.1k & 20.1 \\
 & \small{T5} & 0.73 & 0.22 & 0.72 & 9.9 & 14.2 & 5.9k & 26k & 9.4k & 21.3 \\
\midrule
\multirow{2}{*}{WikiLingua (es$\rightarrow$en)} & \small{mBART} & 0.55 & 0.03 & 0.19 & 8.8 & 14.0 & 4.7k & 63k & 15k & 29.4 \\
 & \small{mBART+} & 0.58 & 0.03 & 0.21 & 9.1 & 14.5 & 5.9k & 83k & 18k & 32.5 \\
  & \small{mT5-small} & 0.39 & 0.03 & 0.15 & 8.3 & 12.8 & 2.3k & 20.9k & 8.2k & 31.8 \\
 & \small{mT5-base} & 0.52 & 0.04 & 0.23 & 8.7 & 13.7 & 3.5k & 34.4k & 10.3k & 28.7 \\
 & \small{mT5-large} & 0.57 & 0.04 & 0.26 & 8.9 & 14.0 & 4.2k & 44.4k & 11.7k & 30.8 \\
 & \small{mT5-XL} & 0.6 & 0.04 & 0.29 & 9.1 & 14.4 & 5.0k & 57.7k & 13.5k & 34.7 \\
\multirow{2}{*}{WikiLingua (ru$\rightarrow$en)} & \small{mBART} & 0.54 & 0.04 & 0.20 & 8.5 & 13.3 & 2.8k & 28k & 8.7k & 27.3 \\
 & \small{mBART+} & 0.55 & 0.04 & 0.23 & 8.8 & 13.7 & 3.5k & 35k & 10k & 28.4  \\
  & \small{mT5-small} & 0.4 & 0.04 & 0.19 & 8.2 & 12.6 & 1.5k & 11.6k & 5.5k & 31.8 \\
 & \small{mT5-base} & 0.55 & 0.06 & 0.3 & 8.6 & 13.4 & 2.5k & 21.0k & 7.1k & 28.7 \\
 & \small{mT5-large} & 0.59 & 0.06 & 0.32 & 8.7 & 13.6 & 3.0k & 26.1k & 7.9k & 31.1 \\
 & \small{mT5-XL} & 0.6 & 0.07 & 0.35 & 8.8 & 13.8 & 3.4k & 29.0k & 8.5k & 31.4 \\
\multirow{2}{*}{WikiLingua (tr$\rightarrow$en)} & \small{mBART} & 0.45 & 0.08 & 0.28 & 7.7 & 11.2 & 743 & 4.1k & 2.1k & 34.2 \\
 & \small{mBART+} & 0.52 & 0.12 & 0.38 & 8.0 & 11.9 & 1.2k & 6.1k & 2.8k & 30.7 \\
  & \small{mT5-small} & 0.55 & 0.14 & 0.46 & 8.1 & 11.6 & 935 & 4.4k & 2.1k & 40.2 \\
 & \small{mT5-base} & 0.59 & 0.16 & 0.51 & 8.2 & 11.9 & 1.0k & 4.8k & 2.2k & 38.7 \\
 & \small{mT5-large} & 0.58 & 0.16 & 0.5 & 8.1 & 11.8 & 1.0k & 4.7k & 2.2k & 38.0 \\
 & \small{mT5-XL} & 0.58 & 0.16 & 0.51 & 8.2 & 11.8 & 1.0k & 4.7k & 2.1k & 36.8 \\
\multirow{2}{*}{WikiLingua (vi$\rightarrow$en)} & \small{mBART} & 0.54 & 0.07 & 0.28 & 8.2 & 12.3 & 1.5k & 9.3k & 4.0k & 26.9  \\
 & \small{mBART+} & 0.54 & 0.08 & 0.33 & 8.6 & 12.9 & 2.1k  & 13k & 5.3k & 29.8 \\
 & \small{mT5-small} & 0.5 & 0.09 & 0.33 & 8.2 & 12.1 & 1.2k & 6.4k & 3.1k & 32.9 \\
 & \small{mT5-base} & 0.58 & 0.12 & 0.43 & 8.4 & 12.6 & 1.6k & 8.9k & 3.7k & 31.1 \\
 & \small{mT5-large} & 0.6 & 0.12 & 0.45 & 8.5 & 12.7 & 1.7k & 9.3k & 3.8k & 30.7 \\
 & \small{mT5-XL} & 0.61 & 0.12 & 0.47 & 8.6 & 12.9 & 1.8k & 10.2k & 4.0k & 31.5 \\
\bottomrule
\end{tabular}
\caption{Results of the baseline results we release with \GEM, focusing on diversity of the outputs and neutral system characterizations.}
\label{tab:result2}
\end{table*}

\paragraph{Probing for Faithfulness.} Another approach that has shown promise in summarization. The approach relies on the insight that a reader of a reference and generated summary should be able to answer the same question, regardless of how the summary is phrased. There has been much development toward these QA-based approaches~\citep[][among others]{eyal2019question,scialom-2019-answers,durmus-2020-feqa,wang-2020-asking} and they can provide an alternative angle to model evaluation that does not highly correlate with other evaluation approaches~\citep{fabbri2020summeval}. While most related work on these metrics is limited to summarization, we are evaluating systems using a QA-based method called QuestEval~\citep{scialom2020SAFEval} that supports all of our tasks. 

In addition to QA-based evaluation, there have also been related efforts to develop more fine-grained and interpretable evaluation metrics, for example to measure consistency in data-to-text problems~\citep{opitz2020towards,dhingra2019handling}. We are using one such metric called NUBIA~\citep{kane-etal-2020-nubia}, the NeUral Based Interchangeability Assessor, which combines multiple measures such as entailment and similarity into a decomposable and interpretable score.

\paragraph{Diversity.} As argued by \citet{hashimoto2019unifying} among many others, NLG models intrinsically trade off diversity and quality. A model can produce more diverse outputs through sampling but at the cost of output quality. To account for this aspect, we compute multiple diversity metrics, starting with those proposed for the analysis of the results of the E2E NLG challenge~\citep{duvsek2020evaluating} and by \citet{vanmiltenburg2018measuring}. These include the Shannon Entropy~\citep{shannon2001mathematical} over unigrams and bigrams ($H_1$, $H_2$), the mean segmented type token ratio over segment lengths of 100~\citep[MSTTR,][]{johnson1944studies}, the ratio of distinct n-grams over the total number of n-grams (Distinct$_{1,2}$), and the count of n-grams that only appear once across the entire test output \citep[Unique$_{1,2}$,][]{li2016diversity}. 

\paragraph{System Characterization.} The final section of metrics will characterize the systems. While the focus of this section will be on qualitative descriptions through model cards, we also gather quantitative information that is not necessarily associated with a judgment. As part of this, we collect the number of parameters of a system, as suggested by \citet{ethayarajh2020utility}. For each task, we additionally report the vocabulary size over the output ($|\mathcal{V}|$) and the mean output length of a system~\citep{sun2019compare}.

\section{Results}

One of the central aims of \GEM{} is to measure the progress in NLG without misrepresenting the complex interactions between the sometimes contradicting measures. We thus will not distill the complex interplay of the data, metrics, and model outputs into a single number or statement, and we do not present results in a traditional leaderboard. Instead, we developed an interactive result exploration system that allows analyses of model results, and which we describe in this section. To further motivate this change, consider the following conclusion someone may draw from looking at a leaderboard:

\begin{displayquote}
System Foo performs the best.
\end{displayquote}

\noindent Our interactive system aims to enable more nuanced statements such as:

\begin{displayquote}
System Foo leads to consistent performance increases in Bar-type metrics on challenges that measure Baz while maintaining equal performance on most metrics of type Qux.
\end{displayquote}

\noindent A screenshot of our system is presented in Figure~\ref{fig:results}.\footnote{An initial version showcasing our baseline results is deployed on our website.} In addition, our baseline results are presented in a tabular view in Tables~\ref{tab:results} and~\ref{tab:result2}. 
Our interactive system is centered around a parallel coordinates plot~\citep{inselberg1985plane} which shows all results as lines through parallel axes. Every line intersects the axes at the corresponding mapped value. For instance, see the red line representing the results for task ``ToTTo'' of baseline ``t5-small''. Filters can be applied along axes (see BLEURT axis in Figure~\ref{fig:results}) and the filtered selection is highlighted through bold lines. A selection can be a set of metrics, systems, or tasks. This style of presentation has not been used before for a benchmark. The closest prior work is by \citet{fu-2020-interpretable} for named-entity recognition which allows similar filtering and sorting, but presents the results in a table. 

However, the parallel coordinates approach can scale to a much greater number of metrics than a table. Moreover, by using a parallel coordinates plot instead of a table, it is easy to spot patterns that span multiple metrics, systems, or tasks. 
For example, the highlighted line in Figure~\ref{fig:results} uncovers that, for the T5 baseline on ToTTo, the diversity metrics score higher than other systems while scoring lower on reference-based metrics. Since we only have a single baseline for ToTTo, it is unclear whether this difference can be attributed to the dataset or the system but this relationship will be uncovered once we receive submissions.

The final system will additionally be able to display the model cards and other related meta-information associated with submissions. It will also be able to show (and compare) exemplary outputs for each test set. Those two features will improve the transparency of the results and systems to those who are not familiar with a task and provide necessary information to those who consider using a particular system.
The combination of all components will enable analysis on quantitative, individual, and qualitative level which can support formulating new research hypotheses and gather in-depth insights about system performance. For example, the functionality to compare human annotation and automatic measures could lead to a better understanding how fluency affect BERTScore.

In addition to the interactive self-directed result exploration, our shared task features an evaluation and analysis part. Instead of dictating the interpretation of the modeling shared task results, we will release all system outputs and metrics in this second part and participants of this part may run their own evaluation and conduct interesting analyses.


\section{Submitting to the benchmark}

While we ask submitters to try to cover as many tasks as possible, we acknowledge potential restrictions on computation resources. We thus do not require that a submissions to \GEM{} has to include predictions on every included test and challenge sets. 
All predictions from a model should be formatted and added into a single file as outlined on our website. 

In addition, we require every submitter to answer a series of questions that we will use to construct a model card~\citep{mitchell2019model} and externalize potential concerns regarding the social impact of a model and its use, or its training data. The card will additionally display information to replicate the experiments. 
While we require responses to these questions at submission time, we allow the information about a model to remain anonymous during required anonymization periods should a paper describing the model be under submission elsewhere. 
All submitted model outputs will be made publicly available for download.  

After a submission, we will run the evaluation suite on the submitted outputs and additionally collect human annotations.

\paragraph{Human Evaluation} \GEM{} will be used to develop reproducible and consistent human evaluation strategies for generated text. This task involves selecting and defining which quantities of the generated text should be measured, developing annotation schemes and rater guidelines to capture these quantities accurately, and infrastructure to annotate system outputs.

We aim to develop these setups for all task setups such as summarization, dialogue, simplification, and data-to-text. To approach this task, we will follow the recently proposed taxonomy of human evaluation measures by \citet{belz2020disentangling} and follow the reporting strategies proposed by \citet{howcroft2020twenty}. The detailed setups will be described in a evaluation datasheet~\citep{shimorina2021human}.  

All shared task participants will be asked to provide gold annotations on system outputs, which we will then use to evaluate the consistency of crowdsourced annotations.\footnote{This approach has been successfully used by WMT for many years. See, e.g., \url{http://www.statmt.org/wmt20/translation-task.html}.}

\section{Next Steps}

This section lists the currently active developments and the long-term steps we will take to ensure that \GEM{} will continue to evolve and improve.

\subsection{Collecting more multilingual data}

Many of the initial datasets in \GEM{} are focused on (American or British) English; we see this release as a starting point for the collection of new datasets to improve the inclusiveness of other languages and cultures. From the task point of view, to ensure the longevity of the dataset, we want it to be practical and socially beneficial. Through \GEM{}, we have developed a set of desired criteria for NLG datasets and we aim to apply this knowledge to data collection and actively work toward reducing the disparity in data availability between languages~\citep{joshi2020state}. To this end, we are focusing on a task that requires content selection, planning, and surface realization along in a grounded scenario. The idea is in the prototyping stage with prospects broadly towards dialog response generation and topic summarization in multiple languages. We plan to do so by collaborating with speakers of low-resourced languages through a participatory research approach, as suggested by~\citep{nekoto2020participatory}. Toward this goal, \GEM{} welcomes anyone interested in collaborating on this effort.

\subsection{Personalizing and Controlling NLG}
\GEM{} currently focuses on tasks that deterministically transform an input into an output. With the increasing use of NLG models in real-world applications, how to enable and evaluate personalized NLG systems (e.g., in dialect or formality) remains challenging. Several related tasks have been proposed, for example, the transfer of writing style from informal to formal~\citep{rao2018dear}, personalization of machine translation systems to align with particular personal traits~\citep{mirkin2015personalized}, or  persona-guided response generation of dialogue systems~\citep{zhang2018personalizing}. We envision our framework to be extended (e.g., dataset, evaluation) to incorporate this line of user-focused NLG.

\subsection{Regular updates to the living benchmark}

To activate the benefits of a living benchmark that is focused on evaluation, we commit to regular updates for \GEM. We invite contributions in the form of model outputs, analyses, and metrics at any time and will automatically update the results presented on our website to incorporate them. 
For the updates to the dataset selection, we want to consider the input of the wider NLG research community. To do so, we will set up a yearly selection process similar to the one described in Section~\ref{sec:selection}. The first update process will be run after the GEM workshop at ACL 2021. 
To be able to have a robust comparison between different versions of \GEM, we will only replace a small subset of datasets at a time. 

\section{Conclusion}

In this paper, we have introduced \GEM, a living natural language generation benchmark with a focus on evaluation. While \GEM{} does not claim to instantly solve all issues of benchmarks in NLG, we aim to provide an environment in which systems can be tested in a principled manner and which can elevate the prominence of interesting evaluation approaches. By providing a testbed to easily conduct experiments across many datasets and evaluate in a repeatable, consistent, and more interpretable way, we will be able to track progress toward the goals in NLG research much more clearly. Moreover, we will be able to extend and shape \GEM{} in the future to include more multilingual datasets, which will assist in their adoption across the wider research community. 

\section{Contribution Statements}

\GEM{} is a large effort with a decentralized organization that is split into different task-specific subgroups. To acknowledge everyone's contribution, we list the contribution statements below for all groups.

\paragraph{Steering Committee.} Antoine Bosselut, Esin Durmus, Varun Prashant Gangal, Sebastian Gehrmann, Laura Perez-Beltrachini, Samira Shaikh, and Wei Xu make up the steering committee. Sebastian Gehrmann coordinates and leads the GEM effort. All others provide feedback and discuss larger decisions regarding the direction of GEM and act as conference organizers for the ACL 2021 workshop. 

\paragraph{Summarization.}
The summarization group members are Chris Emezue, Esin Durmus, Faisal Ladhak, Jiawei Zhou, Juan Diego Rodriguez, Kaustubh Dhole, Khyathi Chandu, Laura Perez, Pawan Sasanka Ammanamanchi, Pedro Henrique Martins, Rubungo Andre Niyongabo, Shashi Narayan, Vikas Raunak, and Yufang Hou.
Pedro Henrique Martins organized the group and wrote the data statement for the MLSum dataset. Pawan Sasanka Ammanamanchi was responsible for the XSum data statement, while Vikas Raunak worked on the Wikilingua statement. Shashi Narayan prepared the GEM version of the XSum dataset and trained its baseline models. Juan Diego Rodriguez was responsible for cleaning the MLSum dataset and trained its baseline models. Faisal Ladhak was responsible for the Wikilingua baseline models. Rubungo Andre Niyongabo participated in the discussions and added related papers to the planning document.

\paragraph{Dialog.} Sashank Santhanam, Samira Shaikh, Bodhisattwa Prasad Majumder, Harsh Jhamtani, Yangfeng Ji, Tosin Adewumi, and Wanyu Du are part of this group. Tosin Adewumi contributed code for DialoGPT, and Wanyu Du trained baselines for Schema-Guided Dialog. Harsh Jhamtani wrote the data card for Wizards of Wikipedia.

\paragraph{Data2Text.} Ondrej Dusek wrote the data cards for E2E NLG and Czech Restaurants
data and a TF loader for Czech Restaurants. He also supplied baseline outputs for E2E, Czech Restaurants, and WebNLG. Sebastian Gehrmann supplied baseline outputs for E2E, WebNLG, and CommonGen. Yacine Jernite wrote the data card for CommonGen and the Hugging Face loaders for Czech Restaurants and WebNLG. Teven Le Scao wrote the Hugging Face loader for E2E. Simon Mille and Anastasia Shimorina wrote the data card for WebNLG.

\paragraph{Table2Text.} Varun Gangal and Miruna Clinciu are part of this group. Miruna Clinciu was responsible primarily for DART and Varun Gangal for ToTTo while maintaining a close correspondence and understanding between them to ensure all steps, such as code structure, preprocessing primitives, baselines were as uniform as possible. 

\paragraph{Simplification.} Dhruv Kumar, Mounica Maddela, and Wei Xu contributed to the GEM Simplification task. Dhruv Kumar created the data cards for the datasets, added Wiki-Auto and Turk/ASSET datasets to TFDS, and integrated the SARI metric \cite{xu-etal-2016-optimizing} into the GEM evaluation framework. Mounica Maddela created baselines for the task and added the Turk benchmark corpus to Hugging Face and TFDS. Wei Xu helped in the organization and planning of the task setup. 

\paragraph{Automated Evaluation.} Ondrej Dusek wrote the base code and included BLEU, Meteor, ROUGE, and referenceless metrics (the latter based on code supplied by Emiel van
Miltenburg). He also prepared reference sets for E2E, Czech
Restaurants and WebNLG. Sebastian Gehrman included BLEURT and BERTScore and prepared the reference sets. Dhruv Kumar included SARI and adapted the code for source-based metrics. Nishant Subramani helped with code refactoring. Miruna Clinciu , Emiel van Miltenburg and Thibault Sellam provided feedback and participated in discussions.

\paragraph{Human Evaluation.} Samira Shaikh was the point of contact for this working group. She led the discussions to make progress on the group goals. She also worked with the group to select the general evaluation criteria as well as the criteria for dialogue and simplification tasks. 
Khyathi Chandu and Miruna Clinciu worked on selecting evaluation criteria for the summarization task and participated in the group discussions.
Simon Mille provided support on using the criteria taxonomy and the annotated evaluation sheets for selecting and defining the criteria to use; worked on selecting the D2T criteria.
Vitaly Nikolaev and Sashank Santhanam  worked on selecting evaluation criteria for dialog and simplification tasks.
João Sedoc worked with the group to select the evaluation criteria in general as well as the specific ones for dialog and simplification. He also helped to select among annotation interfaces.
Anastasia Shimorina worked with the group to select the evaluation criteria and participated in the discussions.
Chris Emezue, Sebastian Gehrmann, Khyati Mahajan, and Yufang Hou participated in discussions.

\paragraph{Website and Submission System.} Aman Madaan, Moin Nadeem, Hendrik Strobelt, and Sebastian Gehrmann are part of this group. Sebastian Gehrmann developed the website. Aman Madaan wrote the initial version of the result presentation. Hendrik Strobelt leads the visualization effort for interactive exploration of results.


\paragraph{Model Infrastructure.} Yacine Jernite wrote the initial script template for evaluating and fine-tuning Hugging Face models with the CommonGen example. Sebastian Gehrmann generalized the script to work with other datasets.
Tosin Adewumi wrote a script for fine-tuning the DialoGPT model for dialogue datasets. Juan Diego Rodriguez worked on the infrastructure to fine-tune mBART on MLSum. Mihir Kale trained all mT5 baselines.

\paragraph{Data and Model Sheets and Statements.} Salomey Osei, Pawan Sasanka Ammanamanchi, Juan Diego Rodriguez, Sebastian Gehrmann, Yacine Jernite, and Angelina McMillan-Major are part of this group. The Data Sheet structure was adapted from a combination of designs created for the Hugging Face Datasets library by Angelina McMillan-Major and Yacine Jernite and one written by Sebastian Gehrmann. Juan Diego Rodriguez and Yacine Jernite wrote initial statements for ASSET and CommonGen respectively. The feedback on those was used to improve the structure of the final template. Everyone contributed to the model card template.

\paragraph{Challenge Sets.} Simon Mille, Emiel van Miltenburg, Kaustubh Dhole, Varun Prashant Gangal, Saad Mahamood, and Laura Perez-Beltrachini proposed and discussed ideas of interest for the data-to-text and the text-to-text tasks. Simon Mille coordinated the group. Emiel van Miltenburg, Saad Mahamood, and Simon Mille worked on the creation of the data-to-text datasets, while Varun Prashant Gangal, Kaustubh Dhole and Laura Perez-Beltrachini worked on the text-to-text datasets. Sebastian Gehrmann contributed the ToTTo challenge set.

\paragraph{Crowdsourcing New Data.} Chris Emezue, Rubungo Andre Niyongabo, Aremu Anuoluwapo, Khyathi Chandu, Yufang Hou, Samira Shaikh, Varun Prashant Gangal, and Dimitra Gkatzia are members of this group. Khyathi Chandu worked on identifying where the current datasets fall short to motivate the crowdsourcing of data for a new task. Based on the suggestions from collaborators, she wrote two task proposals in the domains of long-form text, conversations, and data-to-text that address an array of challenges in generation and easily scale to multiple languages. Samira Shaikh participated in the discussions and gave feedback on the task proposals in the pilot study phase. Dimitra Gkatzia looked into potential resources for crowdsourcing. Chris Emezue and Rubungo Andre Niyongabo explored potential low-resource African languages for crowdsourcing.  We are in the process of piloting the tasks internally.

\vspace{1em}
\noindent The authors of this paper not named in the groups participated in initial discussions, participated in the surveys, and provided regular feedback and guidance. Many participants commented on and helped write this paper. We additionally thank all participants of INLG 2019, the Generation Birds-of-a-Feather meeting at ACL 2020, the EvalNLGEval Workshop at INLG 2020, and members of the generation challenge mailing list of SIGGEN for their participation in the discussions that inspired and influenced the creation of \GEM.

\bibliography{tacl2018}

\begin{thebibliography}{132}
\expandafter\ifx\csname natexlab\endcsname\relax\def\natexlab#1{#1}\fi

\bibitem[{Akoury et~al.(2020)Akoury, Wang, Whiting, Hood, Peng, and
  Iyyer}]{akoury-etal-2020-storium}
Nader Akoury, Shufan Wang, Josh Whiting, Stephen Hood, Nanyun Peng, and Mohit
  Iyyer. 2020.
\newblock \href {https://doi.org/10.18653/v1/2020.emnlp-main.525} {{STORIUM}:
  {A} {D}ataset and {E}valuation {P}latform for {M}achine-in-the-{L}oop {S}tory
  {G}eneration}.
\newblock In \emph{Proceedings of the 2020 Conference on Empirical Methods in
  Natural Language Processing (EMNLP)}, pages 6470--6484, Online. Association
  for Computational Linguistics.

\bibitem[{Alva-Manchego et~al.(2020)Alva-Manchego, Martin, Bordes, Scarton,
  Sagot, and Specia}]{alva-manchego-etal-2020-asset}
Fernando Alva-Manchego, Louis Martin, Antoine Bordes, Carolina Scarton,
  Beno{\^\i}t Sagot, and Lucia Specia. 2020.
\newblock \href {https://doi.org/10.18653/v1/2020.acl-main.424} {{ASSET}: {A}
  dataset for tuning and evaluation of sentence simplification models with
  multiple rewriting transformations}.
\newblock In \emph{Proceedings of the 58th Annual Meeting of the Association
  for Computational Linguistics}, pages 4668--4679, Online. Association for
  Computational Linguistics.

\bibitem[{Anastasopoulos and Neubig(2020)}]{anastasopoulos2020cross}
Antonios Anastasopoulos and Graham Neubig. 2020.
\newblock \href {https://doi.org/10.18653/v1/2020.acl-main.766} {Should all
  cross-lingual embeddings speak {E}nglish?}
\newblock In \emph{Proceedings of the 58th Annual Meeting of the Association
  for Computational Linguistics}, pages 8658--8679, Online. Association for
  Computational Linguistics.

\bibitem[{Bahdanau et~al.(2015)Bahdanau, Cho, and Bengio}]{bahdanau2014neural}
Dzmitry Bahdanau, Kyunghyun Cho, and Yoshua Bengio. 2015.
\newblock \href {http://arxiv.org/abs/1409.0473} {Neural machine translation by
  jointly learning to align and translate}.
\newblock In \emph{3rd International Conference on Learning Representations,
  {ICLR} 2015, San Diego, CA, USA, May 7-9, 2015, Conference Track
  Proceedings}.

\bibitem[{Banerjee and Lavie(2005)}]{banerjee2005meteor}
Satanjeev Banerjee and Alon Lavie. 2005.
\newblock \href {https://www.aclweb.org/anthology/W05-0909/} {{METEOR:} an
  automatic metric for {MT} evaluation with improved correlation with human
  judgments}.
\newblock In \emph{Proceedings of the Workshop on Intrinsic and Extrinsic
  Evaluation Measures for Machine Translation and/or Summarization@ACL 2005,
  Ann Arbor, Michigan, USA, June 29, 2005}, pages 65--72. Association for
  Computational Linguistics.

\bibitem[{Belz et~al.(2011)Belz, White, Espinosa, Kow, Hogan, and
  Stent}]{belz-etal-2011-first}
Anja Belz, Mike White, Dominic Espinosa, Eric Kow, Deirdre Hogan, and Amanda
  Stent. 2011.
\newblock \href {https://www.aclweb.org/anthology/W11-2832} {The first surface
  realisation shared task: Overview and evaluation results}.
\newblock In \emph{Proceedings of the 13th {E}uropean Workshop on Natural
  Language Generation}, pages 217--226, Nancy, France. Association for
  Computational Linguistics.

\bibitem[{Belz et~al.(2020)Belz, Mille, and Howcroft}]{belz2020disentangling}
Anya Belz, Simon Mille, and David~M. Howcroft. 2020.
\newblock \href {https://www.aclweb.org/anthology/2020.inlg-1.24}
  {Disentangling the properties of human evaluation methods: A classification
  system to support comparability, meta-evaluation and reproducibility
  testing}.
\newblock In \emph{Proceedings of the 13th International Conference on Natural
  Language Generation}, pages 183--194, Dublin, Ireland. Association for
  Computational Linguistics.

\bibitem[{Bender(2019)}]{bender2019benderrule}
Emily Bender. 2019.
\newblock \href
  {https://thegradient.pub/the-benderrule-on-naming-the-languages-we-study-and-why-it-matters/}
  {The \#benderrule: On naming the languages we study and why it matters}.
\newblock \emph{The Gradient}.

\bibitem[{Bender(2011)}]{Bender2011OnAA}
Emily~M. Bender. 2011.
\newblock \href
  {http://journals.linguisticsociety.org/elanguage/lilt/article/download/2624/2624-5403-1-PB.pdf}
  {On achieving and evaluating language-independence in {NLP}}.
\newblock \emph{Linguistic Issues in Language Technology}, 6.

\bibitem[{Bender and Friedman(2018)}]{bender2018data}
Emily~M. Bender and Batya Friedman. 2018.
\newblock \href {https://doi.org/10.1162/tacl_a_00041} {Data statements for
  natural language processing: Toward mitigating system bias and enabling
  better science}.
\newblock \emph{Transactions of the Association for Computational Linguistics},
  6:587--604.

\bibitem[{Bojar et~al.(2017)Bojar, Graham, and Kamran}]{bojar2017results}
Ond{\v{r}}ej Bojar, Yvette Graham, and Amir Kamran. 2017.
\newblock \href {https://doi.org/10.18653/v1/W17-4755} {Results of the {WMT}17
  metrics shared task}.
\newblock In \emph{Proceedings of the Second Conference on Machine
  Translation}, pages 489--513, Copenhagen, Denmark. Association for
  Computational Linguistics.

\bibitem[{Bojar et~al.(2016)Bojar, Graham, Kamran, and
  Stanojevi{\'c}}]{bojar2016results}
Ond{\v{r}}ej Bojar, Yvette Graham, Amir Kamran, and Milo{\v{s}} Stanojevi{\'c}.
  2016.
\newblock \href {https://doi.org/10.18653/v1/W16-2302} {Results of the {WMT}16
  metrics shared task}.
\newblock In \emph{Proceedings of the First Conference on Machine Translation:
  Volume 2, Shared Task Papers}, pages 199--231, Berlin, Germany. Association
  for Computational Linguistics.

\bibitem[{Celikyilmaz et~al.(2020)Celikyilmaz, Clark, and
  Gao}]{celikyilmaz2020evaluation}
Asli Celikyilmaz, Elizabeth Clark, and Jianfeng Gao. 2020.
\newblock \href {http://arxiv.org/abs/2006.14799} {Evaluation of text
  generation: {A} survey}.
\newblock \emph{CoRR}, abs/2006.14799.

\bibitem[{Cohan et~al.(2018)Cohan, Dernoncourt, Kim, Bui, Kim, Chang, and
  Goharian}]{cohan2018discourse}
Arman Cohan, Franck Dernoncourt, Doo~Soon Kim, Trung Bui, Seokhwan Kim, Walter
  Chang, and Nazli Goharian. 2018.
\newblock \href {https://doi.org/10.18653/v1/N18-2097} {A discourse-aware
  attention model for abstractive summarization of long documents}.
\newblock In \emph{Proceedings of the 2018 Conference of the North {A}merican
  Chapter of the Association for Computational Linguistics: Human Language
  Technologies, Volume 2 (Short Papers)}, pages 615--621, New Orleans,
  Louisiana. Association for Computational Linguistics.

\bibitem[{Covington et~al.(2006)Covington, He, Brown, Naci, and
  Brown}]{covington2006complex}
Michael~A Covington, Congzhou He, Cati Brown, Lorina Naci, and John Brown.
  2006.
\newblock How complex is that sentence? a proposed revision of the rosenberg
  and abbeduto d-level scale.

\bibitem[{Denton et~al.(2020)Denton, Hanna, Amironesei, Smart, Nicole, and
  Scheuerman}]{denton2020bringing}
Emily Denton, Alex Hanna, Razvan Amironesei, Andrew Smart, Hilary Nicole, and
  Morgan~Klaus Scheuerman. 2020.
\newblock \href {http://arxiv.org/abs/2007.07399} {Bringing the people back in:
  Contesting benchmark machine learning datasets}.
\newblock \emph{CoRR}, abs/2007.07399.

\bibitem[{Devlin et~al.(2019)Devlin, Chang, Lee, and
  Toutanova}]{devlin-2019-bert}
Jacob Devlin, Ming-Wei Chang, Kenton Lee, and Kristina Toutanova. 2019.
\newblock \href {https://doi.org/10.18653/v1/N19-1423} {{BERT}: Pre-training of
  deep bidirectional transformers for language understanding}.
\newblock In \emph{Proceedings of the 2019 Conference of the North {A}merican
  Chapter of the Association for Computational Linguistics: Human Language
  Technologies, Volume 1 (Long and Short Papers)}, pages 4171--4186,
  Minneapolis, Minnesota. Association for Computational Linguistics.

\bibitem[{Dhingra et~al.(2019)Dhingra, Faruqui, Parikh, Chang, Das, and
  Cohen}]{dhingra2019handling}
Bhuwan Dhingra, Manaal Faruqui, Ankur Parikh, Ming-Wei Chang, Dipanjan Das, and
  William Cohen. 2019.
\newblock \href {https://doi.org/10.18653/v1/P19-1483} {Handling divergent
  reference texts when evaluating table-to-text generation}.
\newblock In \emph{Proceedings of the 57th Annual Meeting of the Association
  for Computational Linguistics}, pages 4884--4895, Florence, Italy.
  Association for Computational Linguistics.

\bibitem[{Dieter et~al.(2019)Dieter, Wang, Chaganty, Angeli, and
  Chang}]{dieter-etal-2019-mimic}
Justin Dieter, Tian Wang, Arun~Tejasvi Chaganty, Gabor Angeli, and Angel~X.
  Chang. 2019.
\newblock \href {https://doi.org/10.18653/v1/K19-1037} {Mimic and rephrase:
  Reflective listening in open-ended dialogue}.
\newblock In \emph{Proceedings of the 23rd Conference on Computational Natural
  Language Learning (CoNLL)}, pages 393--403, Hong Kong, China. Association for
  Computational Linguistics.

\bibitem[{Dinan et~al.(2019)Dinan, Roller, Shuster, Fan, Auli, and
  Weston}]{dinan2018wizard}
Emily Dinan, Stephen Roller, Kurt Shuster, Angela Fan, Michael Auli, and Jason
  Weston. 2019.
\newblock \href {https://openreview.net/forum?id=r1l73iRqKm} {Wizard of
  wikipedia: Knowledge-powered conversational agents}.
\newblock In \emph{7th International Conference on Learning Representations,
  {ICLR} 2019, New Orleans, LA, USA, May 6-9, 2019}. OpenReview.net.

\bibitem[{Du et~al.(2017)Du, Shao, and Cardie}]{du2017learning}
Xinya Du, Junru Shao, and Claire Cardie. 2017.
\newblock \href {https://doi.org/10.18653/v1/P17-1123} {Learning to ask: Neural
  question generation for reading comprehension}.
\newblock In \emph{Proceedings of the 55th Annual Meeting of the Association
  for Computational Linguistics (Volume 1: Long Papers)}, pages 1342--1352,
  Vancouver, Canada. Association for Computational Linguistics.

\bibitem[{Dua et~al.(2019)Dua, Gottumukkala, Talmor, Singh, and
  Gardner}]{dua2019orb}
Dheeru Dua, Ananth Gottumukkala, Alon Talmor, Sameer Singh, and Matt Gardner.
  2019.
\newblock \href {https://arxiv.org/abs/1912.12598} {{ORB}: An open reading
  benchmark for comprehensive evaluation of machine reading comprehension}.
\newblock In \emph{EMNLP 2019 MRQA Workshop}, page 147.

\bibitem[{Durmus et~al.(2020)Durmus, He, and Diab}]{durmus-2020-feqa}
Esin Durmus, He~He, and Mona Diab. 2020.
\newblock \href {https://doi.org/10.18653/v1/2020.acl-main.454} {{FEQA}: A
  question answering evaluation framework for faithfulness assessment in
  abstractive summarization}.
\newblock In \emph{Proceedings of the 58th Annual Meeting of the Association
  for Computational Linguistics}, pages 5055--5070, Online. Association for
  Computational Linguistics.

\bibitem[{Du{\v{s}}ek et~al.(2019)Du{\v{s}}ek, Howcroft, and
  Rieser}]{duvsek2019semantic}
Ond{\v{r}}ej Du{\v{s}}ek, David~M. Howcroft, and Verena Rieser. 2019.
\newblock \href {https://doi.org/10.18653/v1/W19-8652} {Semantic noise matters
  for neural natural language generation}.
\newblock In \emph{Proceedings of the 12th International Conference on Natural
  Language Generation}, pages 421--426, Tokyo, Japan. Association for
  Computational Linguistics.

\bibitem[{Du{\v{s}}ek and Jurc{\i}cek(2016)}]{duvsek28context}
Ondrej Du{\v{s}}ek and Filip Jurc{\i}cek. 2016.
\newblock \href {https://ufal.mff.cuni.cz/~odusek/2016/docs/slides.pdf} {A
  context-aware natural language generation dataset for dialogue systems}.
\newblock In \emph{RE-WOCHAT: Workshop on Collecting and Generating Resources
  for Chatbots and Conversational Agents-Development and Evaluation Workshop
  Programme (May 28 th, 2016)}, page~6.

\bibitem[{Du{\v{s}}ek and
  Jur{\v{c}}{\'\i}{\v{c}}ek(2016{\natexlab{a}})}]{dusek2016context}
Ond{\v{r}}ej Du{\v{s}}ek and Filip Jur{\v{c}}{\'\i}{\v{c}}ek.
  2016{\natexlab{a}}.
\newblock \href {https://doi.org/10.18653/v1/W16-3622} {A context-aware natural
  language generator for dialogue systems}.
\newblock In \emph{Proceedings of the 17th Annual Meeting of the Special
  Interest Group on Discourse and Dialogue}, pages 185--190, Los Angeles.
  Association for Computational Linguistics.

\bibitem[{Du{\v{s}}ek and
  Jur{\v{c}}{\'\i}{\v{c}}ek(2016{\natexlab{b}})}]{dusek2016sequence}
Ond{\v{r}}ej Du{\v{s}}ek and Filip Jur{\v{c}}{\'\i}{\v{c}}ek.
  2016{\natexlab{b}}.
\newblock \href {https://doi.org/10.18653/v1/P16-2008} {Sequence-to-sequence
  generation for spoken dialogue via deep syntax trees and strings}.
\newblock In \emph{Proceedings of the 54th Annual Meeting of the Association
  for Computational Linguistics (Volume 2: Short Papers)}, pages 45--51,
  Berlin, Germany. Association for Computational Linguistics.

\bibitem[{Du{\v{s}}ek and Jur{\v{c}}{\'\i}{\v{c}}ek(2019)}]{duvsek2019neural}
Ond{\v{r}}ej Du{\v{s}}ek and Filip Jur{\v{c}}{\'\i}{\v{c}}ek. 2019.
\newblock \href {https://doi.org/10.18653/v1/W19-8670} {Neural generation for
  {C}zech: Data and baselines}.
\newblock In \emph{Proceedings of the 12th International Conference on Natural
  Language Generation}, pages 563--574, Tokyo, Japan. Association for
  Computational Linguistics.

\bibitem[{Dusek et~al.(2020)Dusek, Novikova, and Rieser}]{duvsek2020evaluating}
Ondrej Dusek, Jekaterina Novikova, and Verena Rieser. 2020.
\newblock \href {https://doi.org/10.1016/j.csl.2019.06.009} {Evaluating the
  state-of-the-art of end-to-end natural language generation: The {E2E} {NLG}
  challenge}.
\newblock \emph{Comput. Speech Lang.}, 59:123--156.

\bibitem[{Ethayarajh and Jurafsky(2020)}]{ethayarajh2020utility}
Kawin Ethayarajh and Dan Jurafsky. 2020.
\newblock \href {https://doi.org/10.18653/v1/2020.emnlp-main.393} {Utility is
  in the eye of the user: A critique of {NLP} leaderboards}.
\newblock In \emph{Proceedings of the 2020 Conference on Empirical Methods in
  Natural Language Processing (EMNLP)}, pages 4846--4853, Online. Association
  for Computational Linguistics.

\bibitem[{Eyal et~al.(2019)Eyal, Baumel, and Elhadad}]{eyal2019question}
Matan Eyal, Tal Baumel, and Michael Elhadad. 2019.
\newblock \href {https://doi.org/10.18653/v1/N19-1395} {Question answering as
  an automatic evaluation metric for news article summarization}.
\newblock In \emph{Proceedings of the 2019 Conference of the North {A}merican
  Chapter of the Association for Computational Linguistics: Human Language
  Technologies, Volume 1 (Long and Short Papers)}, pages 3938--3948,
  Minneapolis, Minnesota. Association for Computational Linguistics.

\bibitem[{Fabbri et~al.(2020)Fabbri, Kryscinski, McCann, Xiong, Socher, and
  Radev}]{fabbri2020summeval}
Alexander~R. Fabbri, Wojciech Kryscinski, Bryan McCann, Caiming Xiong, Richard
  Socher, and Dragomir~R. Radev. 2020.
\newblock \href {http://arxiv.org/abs/2007.12626} {Summ{E}val: Re-evaluating
  summarization evaluation}.
\newblock \emph{CoRR}, abs/2007.12626.

\bibitem[{Fan et~al.(2019)Fan, Jernite, Perez, Grangier, Weston, and
  Auli}]{fan-etal-2019-eli5}
Angela Fan, Yacine Jernite, Ethan Perez, David Grangier, Jason Weston, and
  Michael Auli. 2019.
\newblock \href {https://doi.org/10.18653/v1/P19-1346} {{ELI}5: Long form
  question answering}.
\newblock In \emph{Proceedings of the 57th Annual Meeting of the Association
  for Computational Linguistics}, pages 3558--3567, Florence, Italy.
  Association for Computational Linguistics.

\bibitem[{Fan et~al.(2018)Fan, Lewis, and Dauphin}]{fan-etal-2018-hierarchical}
Angela Fan, Mike Lewis, and Yann Dauphin. 2018.
\newblock \href {https://doi.org/10.18653/v1/P18-1082} {Hierarchical neural
  story generation}.
\newblock In \emph{Proceedings of the 56th Annual Meeting of the Association
  for Computational Linguistics (Volume 1: Long Papers)}, pages 889--898,
  Melbourne, Australia. Association for Computational Linguistics.

\bibitem[{Ferreira et~al.(2020)Ferreira, Gardent, van~der Lee, Ilinykh, Mille,
  Moussalem, and Shimorina}]{ferreira20202020}
Thiago~Castro Ferreira, Claire Gardent, Chris van~der Lee, Nikolai Ilinykh,
  Simon Mille, Diego Moussalem, and Anastasia Shimorina. 2020.
\newblock \href
  {https://webnlg-challenge.loria.fr/files/2020.webnlg-papers.7.pdf} {The 2020
  bilingual, bi-directional webnlg+ shared task overview and evaluation results
  (webnlg+ 2020)}.
\newblock In \emph{Proceedings of the 3rd WebNLG Workshop on Natural Language
  Generation from the Semantic Web (WebNLG+ 2020), Dublin, Ireland (Virtual).
  Association for Computational Linguistics}.

\bibitem[{{$\forall$} et~al.(2020){$\forall$}, Nekoto, Marivate, Matsila,
  Fasubaa, Fagbohungbe, Akinola, Muhammad, Kabongo~Kabenamualu, Osei, Sackey,
  Niyongabo, Macharm, Ogayo, Ahia, Berhe, Adeyemi, Mokgesi-Selinga, Okegbemi,
  Martinus, Tajudeen, Degila, Ogueji, Siminyu, Kreutzer, Webster, Ali, Abbott,
  Orife, Ezeani, Dangana, Kamper, Elsahar, Duru, Kioko, Espoir, van Biljon,
  Whitenack, Onyefuluchi, Emezue, Dossou, Sibanda, Bassey, Olabiyi, Ramkilowan,
  {\"O}ktem, Akinfaderin, and Bashir}]{nekoto2020participatory}
{ } {$\forall$}, Wilhelmina Nekoto, Vukosi Marivate, Tshinondiwa Matsila, Timi
  Fasubaa, Taiwo Fagbohungbe, Solomon~Oluwole Akinola, Shamsuddeen Muhammad,
  Salomon Kabongo~Kabenamualu, Salomey Osei, Freshia Sackey, Rubungo~Andre
  Niyongabo, Ricky Macharm, Perez Ogayo, Orevaoghene Ahia, Musie~Meressa Berhe,
  Mofetoluwa Adeyemi, Masabata Mokgesi-Selinga, Lawrence Okegbemi, Laura
  Martinus, Kolawole Tajudeen, Kevin Degila, Kelechi Ogueji, Kathleen Siminyu,
  Julia Kreutzer, Jason Webster, Jamiil~Toure Ali, Jade Abbott, Iroro Orife,
  Ignatius Ezeani, Idris~Abdulkadir Dangana, Herman Kamper, Hady Elsahar,
  Goodness Duru, Ghollah Kioko, Murhabazi Espoir, Elan van Biljon, Daniel
  Whitenack, Christopher Onyefuluchi, Chris~Chinenye Emezue, Bonaventure F.~P.
  Dossou, Blessing Sibanda, Blessing Bassey, Ayodele Olabiyi, Arshath
  Ramkilowan, Alp {\"O}ktem, Adewale Akinfaderin, and Abdallah Bashir. 2020.
\newblock \href {https://doi.org/10.18653/v1/2020.findings-emnlp.195}
  {Participatory research for low-resourced machine translation: A case study
  in {A}frican languages}.
\newblock In \emph{Findings of the Association for Computational Linguistics:
  EMNLP 2020}, pages 2144--2160, Online. Association for Computational
  Linguistics.

\bibitem[{Freitag et~al.(2020)Freitag, Grangier, and
  Caswell}]{freitag-2020-bleu}
Markus Freitag, David Grangier, and Isaac Caswell. 2020.
\newblock \href {https://doi.org/10.18653/v1/2020.emnlp-main.5} {{BLEU} might
  be guilty but references are not innocent}.
\newblock In \emph{Proceedings of the 2020 Conference on Empirical Methods in
  Natural Language Processing (EMNLP)}, pages 61--71, Online. Association for
  Computational Linguistics.

\bibitem[{Fu et~al.(2020)Fu, Liu, and Neubig}]{fu-2020-interpretable}
Jinlan Fu, Pengfei Liu, and Graham Neubig. 2020.
\newblock \href {https://doi.org/10.18653/v1/2020.emnlp-main.489}
  {Interpretable multi-dataset evaluation for named entity recognition}.
\newblock In \emph{Proceedings of the 2020 Conference on Empirical Methods in
  Natural Language Processing (EMNLP)}, pages 6058--6069, Online. Association
  for Computational Linguistics.

\bibitem[{Gabriel et~al.(2020)Gabriel, Celikyilmaz, Jha, Choi, and
  Gao}]{gabriel2020go}
Saadia Gabriel, Asli Celikyilmaz, Rahul Jha, Yejin Choi, and Jianfeng Gao.
  2020.
\newblock \href {http://arxiv.org/abs/2010.12834} {Go figure! {A} meta
  evaluation of factuality in summarization}.
\newblock \emph{CoRR}, abs/2010.12834.

\bibitem[{Gardent et~al.(2017)Gardent, Shimorina, Narayan, and
  Perez-Beltrachini}]{gardent2017webnlg}
Claire Gardent, Anastasia Shimorina, Shashi Narayan, and Laura
  Perez-Beltrachini. 2017.
\newblock \href {https://doi.org/10.18653/v1/W17-3518} {The {W}eb{NLG}
  challenge: Generating text from {RDF} data}.
\newblock In \emph{Proceedings of the 10th International Conference on Natural
  Language Generation}, pages 124--133, Santiago de Compostela, Spain.
  Association for Computational Linguistics.

\bibitem[{Gatt and Krahmer(2018)}]{gatt2018survey}
Albert Gatt and Emiel Krahmer. 2018.
\newblock \href {https://doi.org/10.1613/jair.5477} {Survey of the state of the
  art in natural language generation: Core tasks, applications and evaluation}.
\newblock \emph{J. Artif. Intell. Res.}, 61:65--170.

\bibitem[{Gebru et~al.(2018)Gebru, Morgenstern, Vecchione, Vaughan, Wallach,
  Daum{\'e}~III, and Crawford}]{gebru2018datasheets}
Timnit Gebru, Jamie Morgenstern, Briana Vecchione, Jennifer~Wortman Vaughan,
  Hanna Wallach, Hal Daum{\'e}~III, and Kate Crawford. 2018.
\newblock \href {http://arxiv.org/abs/1803.09010} {Datasheets for datasets}.
\newblock In \emph{Proceedings of the Fifth Workshop on Fairness,
  Accountability, and Transparency in Machine Learning}, Stockholm, Sweden.

\bibitem[{Hashimoto et~al.(2019)Hashimoto, Zhang, and
  Liang}]{hashimoto2019unifying}
Tatsunori Hashimoto, Hugh Zhang, and Percy Liang. 2019.
\newblock \href {https://doi.org/10.18653/v1/N19-1169} {Unifying human and
  statistical evaluation for natural language generation}.
\newblock In \emph{Proceedings of the 2019 Conference of the North {A}merican
  Chapter of the Association for Computational Linguistics: Human Language
  Technologies, Volume 1 (Long and Short Papers)}, pages 1689--1701,
  Minneapolis, Minnesota. Association for Computational Linguistics.

\bibitem[{Heafield et~al.(2020)Heafield, Hayashi, Oda, Konstas, Finch, Neubig,
  Li, and Birch}]{heafield2020findings}
Kenneth Heafield, Hiroaki Hayashi, Yusuke Oda, Ioannis Konstas, Andrew Finch,
  Graham Neubig, Xian Li, and Alexandra Birch. 2020.
\newblock \href {https://doi.org/10.18653/v1/2020.ngt-1.1} {Findings of the
  fourth workshop on neural generation and translation}.
\newblock In \emph{Proceedings of the Fourth Workshop on Neural Generation and
  Translation}, pages 1--9, Online. Association for Computational Linguistics.

\bibitem[{Hermann et~al.(2015)Hermann, Kocisk{\'{y}}, Grefenstette, Espeholt,
  Kay, Suleyman, and Blunsom}]{hermann2015teaching}
Karl~Moritz Hermann, Tom{\'{a}}s Kocisk{\'{y}}, Edward Grefenstette, Lasse
  Espeholt, Will Kay, Mustafa Suleyman, and Phil Blunsom. 2015.
\newblock \href
  {https://proceedings.neurips.cc/paper/2015/hash/afdec7005cc9f14302cd0474fd0f3c96-Abstract.html}
  {Teaching machines to read and comprehend}.
\newblock In \emph{Advances in Neural Information Processing Systems 28: Annual
  Conference on Neural Information Processing Systems 2015, December 7-12,
  2015, Montreal, Quebec, Canada}, pages 1693--1701.

\bibitem[{Howcroft et~al.(2020)Howcroft, Belz, Clinciu, Gkatzia, Hasan,
  Mahamood, Mille, van Miltenburg, Santhanam, and Rieser}]{howcroft2020twenty}
David~M. Howcroft, Anya Belz, Miruna-Adriana Clinciu, Dimitra Gkatzia, Sadid~A.
  Hasan, Saad Mahamood, Simon Mille, Emiel van Miltenburg, Sashank Santhanam,
  and Verena Rieser. 2020.
\newblock \href {https://www.aclweb.org/anthology/2020.inlg-1.23} {Twenty years
  of confusion in human evaluation: {NLG} needs evaluation sheets and
  standardised definitions}.
\newblock In \emph{Proceedings of the 13th International Conference on Natural
  Language Generation}, pages 169--182, Dublin, Ireland. Association for
  Computational Linguistics.

\bibitem[{Hu et~al.(2015)Hu, Chen, and Zhu}]{hu-etal-2015-lcsts}
Baotian Hu, Qingcai Chen, and Fangze Zhu. 2015.
\newblock \href {https://doi.org/10.18653/v1/D15-1229} {{LCSTS}: A large scale
  {C}hinese short text summarization dataset}.
\newblock In \emph{Proceedings of the 2015 Conference on Empirical Methods in
  Natural Language Processing}, pages 1967--1972, Lisbon, Portugal. Association
  for Computational Linguistics.

\bibitem[{Hu et~al.(2020)Hu, Ruder, Siddhant, Neubig, Firat, and
  Johnson}]{hu2020xtreme}
Junjie Hu, Sebastian Ruder, Aditya Siddhant, Graham Neubig, Orhan Firat, and
  Melvin Johnson. 2020.
\newblock \href {http://proceedings.mlr.press/v119/hu20b.html} {{XTREME:} {A}
  massively multilingual multi-task benchmark for evaluating cross-lingual
  generalisation}.
\newblock In \emph{Proceedings of the 37th International Conference on Machine
  Learning, {ICML} 2020, 13-18 July 2020, Virtual Event}, volume 119 of
  \emph{Proceedings of Machine Learning Research}, pages 4411--4421. {PMLR}.

\bibitem[{Inselberg(1985)}]{inselberg1985plane}
Alfred Inselberg. 1985.
\newblock \href {https://doi.org/10.1007/BF01898350} {The plane with parallel
  coordinates}.
\newblock \emph{Vis. Comput.}, 1(2):69--91.

\bibitem[{Jiang et~al.(2020)Jiang, Maddela, Lan, Zhong, and
  Xu}]{jiang-etal-2020-neural}
Chao Jiang, Mounica Maddela, Wuwei Lan, Yang Zhong, and Wei Xu. 2020.
\newblock \href {https://doi.org/10.18653/v1/2020.acl-main.709} {Neural {CRF}
  model for sentence alignment in text simplification}.
\newblock In \emph{Proceedings of the 58th Annual Meeting of the Association
  for Computational Linguistics}, pages 7943--7960, Online. Association for
  Computational Linguistics.

\bibitem[{Jishan et~al.(2019)Jishan, Mahmud, and Azad}]{jishan2020bangla}
Md.~Asifuzzaman Jishan, Khan~Raqib Mahmud, and Abul Kalam~Al Azad. 2019.
\newblock \href {https://doi.org/10.7910/DVN/DZZ1ZB} {{Bangla Natural Language
  Image to Text (BNLIT)}}.

\bibitem[{Johnson(1944)}]{johnson1944studies}
Wendell Johnson. 1944.
\newblock Studies in language behavior: A program of research.
\newblock \emph{Psychological Monographs}, 56(2):1--15.

\bibitem[{Joshi et~al.(2020)Joshi, Santy, Budhiraja, Bali, and
  Choudhury}]{joshi2020state}
Pratik Joshi, Sebastin Santy, Amar Budhiraja, Kalika Bali, and Monojit
  Choudhury. 2020.
\newblock \href {https://doi.org/10.18653/v1/2020.acl-main.560} {The state and
  fate of linguistic diversity and inclusion in the {NLP} world}.
\newblock In \emph{Proceedings of the 58th Annual Meeting of the Association
  for Computational Linguistics}, pages 6282--6293, Online. Association for
  Computational Linguistics.

\bibitem[{Kale and Rastogi(2020)}]{kale2020few}
Mihir Kale and Abhinav Rastogi. 2020.
\newblock Few-shot natural language generation by rewriting templates.
\newblock \emph{arXiv preprint arXiv:2004.15006}.

\bibitem[{Kane et~al.(2020)Kane, Kocyigit, Abdalla, Ajanoh, and
  Coulibali}]{kane-etal-2020-nubia}
Hassan Kane, Muhammed~Yusuf Kocyigit, Ali Abdalla, Pelkins Ajanoh, and Mohamed
  Coulibali. 2020.
\newblock \href {https://www.aclweb.org/anthology/2020.evalnlgeval-1.4}
  {{NUBIA}: {N}e{U}ral based interchangeability assessor for text generation}.
\newblock In \emph{Proceedings of the 1st Workshop on Evaluating NLG
  Evaluation}, pages 28--37, Online (Dublin, Ireland). Association for
  Computational Linguistics.

\bibitem[{Kedzie et~al.(2018)Kedzie, McKeown, and
  Daum{\'e}~III}]{kedzie2018content}
Chris Kedzie, Kathleen McKeown, and Hal Daum{\'e}~III. 2018.
\newblock \href {https://doi.org/10.18653/v1/D18-1208} {Content selection in
  deep learning models of summarization}.
\newblock In \emph{Proceedings of the 2018 Conference on Empirical Methods in
  Natural Language Processing}, pages 1818--1828, Brussels, Belgium.
  Association for Computational Linguistics.

\bibitem[{Khashabi et~al.(2021)Khashabi, Stanovsky, Bragg, Lourie, Kasai, Choi,
  Smith, and Weld}]{khasabi2021genie}
Daniel Khashabi, Gabriel Stanovsky, Jonathan Bragg, Nicholas Lourie, Jungo
  Kasai, Yejin Choi, Noah~A. Smith, and Daniel~S. Weld. 2021.
\newblock \href {http://arxiv.org/abs/2101.06561} {{GENIE:} {A} leaderboard for
  human-in-the-loop evaluation of text generation}.
\newblock \emph{CoRR}, abs/2101.06561.

\bibitem[{Ko{\v{c}}isk{\`y} et~al.(2018)Ko{\v{c}}isk{\`y}, Schwarz, Blunsom,
  Dyer, Hermann, Melis, and Grefenstette}]{kovcisky2018narrativeqa}
Tom{\'a}{\v{s}} Ko{\v{c}}isk{\`y}, Jonathan Schwarz, Phil Blunsom, Chris Dyer,
  Karl~Moritz Hermann, G{\'a}bor Melis, and Edward Grefenstette. 2018.
\newblock The narrative{QA} reading comprehension challenge.
\newblock \emph{Transactions of the Association for Computational Linguistics},
  6:317--328.

\bibitem[{Ladhak et~al.(2020)Ladhak, Durmus, Cardie, and
  McKeown}]{ladhak2020wikilingua}
Faisal Ladhak, Esin Durmus, Claire Cardie, and Kathleen McKeown. 2020.
\newblock \href {https://doi.org/10.18653/v1/2020.findings-emnlp.360}
  {{W}iki{L}ingua: A new benchmark dataset for cross-lingual abstractive
  summarization}.
\newblock In \emph{Findings of the Association for Computational Linguistics:
  EMNLP 2020}, pages 4034--4048, Online. Association for Computational
  Linguistics.

\bibitem[{Lebret et~al.(2016)Lebret, Grangier, and Auli}]{lebret2016neural}
R{\'e}mi Lebret, David Grangier, and Michael Auli. 2016.
\newblock \href {https://doi.org/10.18653/v1/D16-1128} {Neural text generation
  from structured data with application to the biography domain}.
\newblock In \emph{Proceedings of the 2016 Conference on Empirical Methods in
  Natural Language Processing}, pages 1203--1213, Austin, Texas. Association
  for Computational Linguistics.

\bibitem[{Lewis et~al.(2020{\natexlab{a}})Lewis, Liu, Goyal, Ghazvininejad,
  Mohamed, Levy, Stoyanov, and Zettlemoyer}]{lewis2019bart}
Mike Lewis, Yinhan Liu, Naman Goyal, Marjan Ghazvininejad, Abdelrahman Mohamed,
  Omer Levy, Veselin Stoyanov, and Luke Zettlemoyer. 2020{\natexlab{a}}.
\newblock \href {https://doi.org/10.18653/v1/2020.acl-main.703} {{BART}:
  Denoising sequence-to-sequence pre-training for natural language generation,
  translation, and comprehension}.
\newblock In \emph{Proceedings of the 58th Annual Meeting of the Association
  for Computational Linguistics}, pages 7871--7880, Online. Association for
  Computational Linguistics.

\bibitem[{Lewis et~al.(2020{\natexlab{b}})Lewis, Oguz, Rinott, Riedel, and
  Schwenk}]{lewis2020mlqa}
Patrick Lewis, Barlas Oguz, Ruty Rinott, Sebastian Riedel, and Holger Schwenk.
  2020{\natexlab{b}}.
\newblock \href {https://doi.org/10.18653/v1/2020.acl-main.653} {{MLQA}:
  Evaluating cross-lingual extractive question answering}.
\newblock In \emph{Proceedings of the 58th Annual Meeting of the Association
  for Computational Linguistics}, pages 7315--7330, Online. Association for
  Computational Linguistics.

\bibitem[{Li et~al.(2016)Li, Galley, Brockett, Gao, and
  Dolan}]{li2016diversity}
Jiwei Li, Michel Galley, Chris Brockett, Jianfeng Gao, and Bill Dolan. 2016.
\newblock \href {https://doi.org/10.18653/v1/N16-1014} {A diversity-promoting
  objective function for neural conversation models}.
\newblock In \emph{Proceedings of the 2016 Conference of the North {A}merican
  Chapter of the Association for Computational Linguistics: Human Language
  Technologies}, pages 110--119, San Diego, California. Association for
  Computational Linguistics.

\bibitem[{Li et~al.(2018)Li, Duan, Zhou, Chu, Ouyang, Wang, and
  Zhou}]{li2018visual}
Yikang Li, Nan Duan, Bolei Zhou, Xiao Chu, Wanli Ouyang, Xiaogang Wang, and
  Ming Zhou. 2018.
\newblock \href {https://doi.org/10.1109/CVPR.2018.00640} {Visual question
  generation as dual task of visual question answering}.
\newblock In \emph{2018 {IEEE} Conference on Computer Vision and Pattern
  Recognition, {CVPR} 2018, Salt Lake City, UT, USA, June 18-22, 2018}, pages
  6116--6124. {IEEE} Computer Society.

\bibitem[{Liang et~al.(2020)Liang, Duan, Gong, Wu, Guo, Qi, Gong, Shou, Jiang,
  Cao, Fan, Zhang, Agrawal, Cui, Wei, Bharti, Qiao, Chen, Wu, Liu, Yang,
  Majumder, and Zhou}]{liang2020xglue}
Yaobo Liang, Nan Duan, Yeyun Gong, Ning Wu, Fenfei Guo, Weizhen Qi, Ming Gong,
  Linjun Shou, Daxin Jiang, Guihong Cao, Xiaodong Fan, Bruce Zhang, Rahul
  Agrawal, Edward Cui, Sining Wei, Taroon Bharti, Ying Qiao, Jiun{-}Hung Chen,
  Winnie Wu, Shuguang Liu, Fan Yang, Rangan Majumder, and Ming Zhou. 2020.
\newblock \href {http://arxiv.org/abs/2004.01401} {{XGLUE:} {A} new benchmark
  dataset for cross-lingual pre-training, understanding and generation}.
\newblock \emph{CoRR}, abs/2004.01401.

\bibitem[{Lin et~al.(2020)Lin, Zhou, Shen, Zhou, Bhagavatula, Choi, and
  Ren}]{lin2019commongen}
Bill~Yuchen Lin, Wangchunshu Zhou, Ming Shen, Pei Zhou, Chandra Bhagavatula,
  Yejin Choi, and Xiang Ren. 2020.
\newblock \href {https://doi.org/10.18653/v1/2020.findings-emnlp.165}
  {{C}ommon{G}en: A constrained text generation challenge for generative
  commonsense reasoning}.
\newblock In \emph{Findings of the Association for Computational Linguistics:
  EMNLP 2020}, pages 1823--1840, Online. Association for Computational
  Linguistics.

\bibitem[{Lin(2004)}]{lin2004rouge}
Chin-Yew Lin. 2004.
\newblock \href {https://www.aclweb.org/anthology/W04-1013} {{ROUGE}: A package
  for automatic evaluation of summaries}.
\newblock In \emph{Text Summarization Branches Out}, pages 74--81, Barcelona,
  Spain. Association for Computational Linguistics.

\bibitem[{Linzen(2020)}]{linzen2020accelerate}
Tal Linzen. 2020.
\newblock \href {https://doi.org/10.18653/v1/2020.acl-main.465} {How can we
  accelerate progress towards human-like linguistic generalization?}
\newblock In \emph{Proceedings of the 58th Annual Meeting of the Association
  for Computational Linguistics}, pages 5210--5217, Online. Association for
  Computational Linguistics.

\bibitem[{Liu et~al.(2020{\natexlab{a}})Liu, Yan, Gong, Qi, Zhang, Jiao, Chen,
  Fu, Shou, Gong, Wang, Chen, Jiang, Lv, Zhang, Wu, Zhou, and
  Duan}]{liu2020glge}
Dayiheng Liu, Yu~Yan, Yeyun Gong, Weizhen Qi, Hang Zhang, Jian Jiao, Weizhu
  Chen, Jie Fu, Linjun Shou, Ming Gong, Pengcheng Wang, Jiusheng Chen, Daxin
  Jiang, Jiancheng Lv, Ruofei Zhang, Winnie Wu, Ming Zhou, and Nan Duan.
  2020{\natexlab{a}}.
\newblock \href {http://arxiv.org/abs/2011.11928} {{GLGE:} {A} new general
  language generation evaluation benchmark}.
\newblock \emph{CoRR}, abs/2011.11928.

\bibitem[{Liu et~al.(2018)Liu, Saleh, Pot, Goodrich, Sepassi, Kaiser, and
  Shazeer}]{liu2018generating}
Peter~J. Liu, Mohammad Saleh, Etienne Pot, Ben Goodrich, Ryan Sepassi, Lukasz
  Kaiser, and Noam Shazeer. 2018.
\newblock \href {https://openreview.net/forum?id=Hyg0vbWC-} {Generating
  wikipedia by summarizing long sequences}.
\newblock In \emph{6th International Conference on Learning Representations,
  {ICLR} 2018, Vancouver, BC, Canada, April 30 - May 3, 2018, Conference Track
  Proceedings}. OpenReview.net.

\bibitem[{Liu et~al.(2020{\natexlab{b}})Liu, Gu, Goyal, Li, Edunov,
  Ghazvininejad, Lewis, and Zettlemoyer}]{liu2020multilingual}
Yinhan Liu, Jiatao Gu, Naman Goyal, Xian Li, Sergey Edunov, Marjan
  Ghazvininejad, Mike Lewis, and Luke Zettlemoyer. 2020{\natexlab{b}}.
\newblock \href {https://transacl.org/ojs/index.php/tacl/article/view/2107}
  {Multilingual denoising pre-training for neural machine translation}.
\newblock \emph{Trans. Assoc. Comput. Linguistics}, 8:726--742.

\bibitem[{Liu et~al.(2019)Liu, Ott, Goyal, Du, Joshi, Chen, Levy, Lewis,
  Zettlemoyer, and Stoyanov}]{liu2019roberta}
Yinhan Liu, Myle Ott, Naman Goyal, Jingfei Du, Mandar Joshi, Danqi Chen, Omer
  Levy, Mike Lewis, Luke Zettlemoyer, and Veselin Stoyanov. 2019.
\newblock \href {http://arxiv.org/abs/1907.11692} {Roberta: {A} robustly
  optimized {BERT} pretraining approach}.
\newblock \emph{CoRR}, abs/1907.11692.

\bibitem[{Lowe et~al.(2015)Lowe, Pow, Serban, and
  Pineau}]{lowe-etal-2015-ubuntu}
Ryan Lowe, Nissan Pow, Iulian Serban, and Joelle Pineau. 2015.
\newblock \href {https://doi.org/10.18653/v1/W15-4640} {The {U}buntu dialogue
  corpus: A large dataset for research in unstructured multi-turn dialogue
  systems}.
\newblock In \emph{Proceedings of the 16th Annual Meeting of the Special
  Interest Group on Discourse and Dialogue}, pages 285--294, Prague, Czech
  Republic. Association for Computational Linguistics.

\bibitem[{Lu(2010)}]{lu2010automatic}
Xiaofei Lu. 2010.
\newblock Automatic analysis of syntactic complexity in second language
  writing.
\newblock \emph{International Journal of Corpus Linguistics}, 15(4):474--496.

\bibitem[{Ma et~al.(2018)Ma, Bojar, and Graham}]{ma2018results}
Qingsong Ma, Ond{\v{r}}ej Bojar, and Yvette Graham. 2018.
\newblock \href {https://doi.org/10.18653/v1/W18-6450} {Results of the {WMT}18
  metrics shared task: Both characters and embeddings achieve good
  performance}.
\newblock In \emph{Proceedings of the Third Conference on Machine Translation:
  Shared Task Papers}, pages 671--688, Belgium, Brussels. Association for
  Computational Linguistics.

\bibitem[{Ma et~al.(2019)Ma, Wei, Bojar, and Graham}]{ma2019results}
Qingsong Ma, Johnny Wei, Ond{\v{r}}ej Bojar, and Yvette Graham. 2019.
\newblock \href {https://doi.org/10.18653/v1/W19-5302} {Results of the {WMT}19
  metrics shared task: Segment-level and strong {MT} systems pose big
  challenges}.
\newblock In \emph{Proceedings of the Fourth Conference on Machine Translation
  (Volume 2: Shared Task Papers, Day 1)}, pages 62--90, Florence, Italy.
  Association for Computational Linguistics.

\bibitem[{Manning et~al.(2020)Manning, Wein, and Schneider}]{manning2020human}
Emma Manning, Shira Wein, and Nathan Schneider. 2020.
\newblock \href {https://doi.org/10.18653/v1/2020.coling-main.420} {A human
  evaluation of amr-to-english generation systems}.
\newblock In \emph{Proceedings of the 28th International Conference on
  Computational Linguistics, {COLING} 2020, Barcelona, Spain (Online), December
  8-13, 2020}, pages 4773--4786. International Committee on Computational
  Linguistics.

\bibitem[{Mathur et~al.(2020{\natexlab{a}})Mathur, Baldwin, and
  Cohn}]{mathur-2020-tangled}
Nitika Mathur, Timothy Baldwin, and Trevor Cohn. 2020{\natexlab{a}}.
\newblock \href {https://doi.org/10.18653/v1/2020.acl-main.448} {Tangled up in
  {BLEU}: Reevaluating the evaluation of automatic machine translation
  evaluation metrics}.
\newblock In \emph{Proceedings of the 58th Annual Meeting of the Association
  for Computational Linguistics}, pages 4984--4997, Online. Association for
  Computational Linguistics.

\bibitem[{Mathur et~al.(2020{\natexlab{b}})Mathur, Wei, Freitag, Ma, and
  Bojar}]{mathur2020results}
Nitika Mathur, Johnny Wei, Markus Freitag, Qingsong Ma, and Ond{\v{r}}ej Bojar.
  2020{\natexlab{b}}.
\newblock \href {https://www.aclweb.org/anthology/2020.wmt-1.77} {Results of
  the {WMT}20 metrics shared task}.
\newblock In \emph{Proceedings of the Fifth Conference on Machine Translation},
  pages 688--725, Online. Association for Computational Linguistics.

\bibitem[{Maynez et~al.(2020)Maynez, Narayan, Bohnet, and
  McDonald}]{maynez2020faithfulness}
Joshua Maynez, Shashi Narayan, Bernd Bohnet, and Ryan McDonald. 2020.
\newblock \href {https://doi.org/10.18653/v1/2020.acl-main.173} {On
  faithfulness and factuality in abstractive summarization}.
\newblock In \emph{Proceedings of the 58th Annual Meeting of the Association
  for Computational Linguistics}, pages 1906--1919, Online. Association for
  Computational Linguistics.

\bibitem[{McCann et~al.(2018)McCann, Keskar, Xiong, and
  Socher}]{mccann2018natural}
Bryan McCann, Nitish~Shirish Keskar, Caiming Xiong, and Richard Socher. 2018.
\newblock \href {http://arxiv.org/abs/1806.08730} {The natural language
  decathlon: Multitask learning as question answering}.
\newblock \emph{CoRR}, abs/1806.08730.

\bibitem[{Mille et~al.(2018)Mille, Belz, Bohnet, Graham, Pitler, and
  Wanner}]{mille2018first}
Simon Mille, Anja Belz, Bernd Bohnet, Yvette Graham, Emily Pitler, and Leo
  Wanner. 2018.
\newblock \href {https://doi.org/10.18653/v1/W18-3601} {The first multilingual
  surface realisation shared task ({SR}{'}18): Overview and evaluation
  results}.
\newblock In \emph{Proceedings of the First Workshop on Multilingual Surface
  Realisation}, pages 1--12, Melbourne, Australia. Association for
  Computational Linguistics.

\bibitem[{Mille et~al.(2019)Mille, Belz, Bohnet, Graham, and
  Wanner}]{mille2019proceedings}
Simon Mille, Anja Belz, Bernd Bohnet, Yvette Graham, and Leo Wanner, editors.
  2019.
\newblock \href {https://www.aclweb.org/anthology/D19-6300} {\emph{Proceedings
  of the 2nd Workshop on Multilingual Surface Realisation (MSR 2019)}}.
  Association for Computational Linguistics, Hong Kong, China.

\bibitem[{Mille et~al.(2020)Mille, Belz, Bohnet, Castro~Ferreira, Graham, and
  Wanner}]{mille2020third}
Simon Mille, Anya Belz, Bernd Bohnet, Thiago Castro~Ferreira, Yvette Graham,
  and Leo Wanner. 2020.
\newblock \href {https://www.aclweb.org/anthology/2020.msr-1.1} {The third
  multilingual surface realisation shared task ({SR}{'}20): Overview and
  evaluation results}.
\newblock In \emph{Proceedings of the Third Workshop on Multilingual Surface
  Realisation}, pages 1--20, Barcelona, Spain (Online). Association for
  Computational Linguistics.

\bibitem[{van Miltenburg et~al.(2018)van Miltenburg, Elliott, and
  Vossen}]{vanmiltenburg2018measuring}
Emiel van Miltenburg, Desmond Elliott, and Piek Vossen. 2018.
\newblock \href {https://www.aclweb.org/anthology/C18-1147} {Measuring the
  diversity of automatic image descriptions}.
\newblock In \emph{Proceedings of the 27th International Conference on
  Computational Linguistics}, pages 1730--1741, Santa Fe, New Mexico, USA.
  Association for Computational Linguistics.

\bibitem[{Min et~al.(2020)Min, Michael, Hajishirzi, and
  Zettlemoyer}]{min2020ambigqa}
Sewon Min, Julian Michael, Hannaneh Hajishirzi, and Luke Zettlemoyer. 2020.
\newblock \href {https://doi.org/10.18653/v1/2020.emnlp-main.466} {{A}mbig{QA}:
  Answering ambiguous open-domain questions}.
\newblock In \emph{Proceedings of the 2020 Conference on Empirical Methods in
  Natural Language Processing (EMNLP)}, pages 5783--5797, Online. Association
  for Computational Linguistics.

\bibitem[{Mirkin and Meunier(2015)}]{mirkin2015personalized}
Shachar Mirkin and Jean-Luc Meunier. 2015.
\newblock \href {https://doi.org/10.18653/v1/D15-1238} {Personalized machine
  translation: Predicting translational preferences}.
\newblock In \emph{Proceedings of the 2015 Conference on Empirical Methods in
  Natural Language Processing}, pages 2019--2025, Lisbon, Portugal. Association
  for Computational Linguistics.

\bibitem[{Mitchell et~al.(2019)Mitchell, Wu, Zaldivar, Barnes, Vasserman,
  Hutchinson, Spitzer, Raji, and Gebru}]{mitchell2019model}
Margaret Mitchell, Simone Wu, Andrew Zaldivar, Parker Barnes, Lucy Vasserman,
  Ben Hutchinson, Elena Spitzer, Inioluwa~Deborah Raji, and Timnit Gebru. 2019.
\newblock Model cards for model reporting.
\newblock In \emph{Proceedings of the conference on fairness, accountability,
  and transparency}, pages 220--229.

\bibitem[{Nallapati et~al.(2016)Nallapati, Zhou, dos Santos,
  GuÌ‡l{\c{c}}ehre, and Xiang}]{nallapati2016abstractive}
Ramesh Nallapati, Bowen Zhou, Cicero dos Santos, {\c{C}}a{\u{g}}lar
  GuÌ‡l{\c{c}}ehre, and Bing Xiang. 2016.
\newblock \href {https://doi.org/10.18653/v1/K16-1028} {Abstractive text
  summarization using sequence-to-sequence {RNN}s and beyond}.
\newblock In \emph{Proceedings of The 20th {SIGNLL} Conference on Computational
  Natural Language Learning}, pages 280--290, Berlin, Germany. Association for
  Computational Linguistics.

\bibitem[{Narayan et~al.(2018)Narayan, Cohen, and Lapata}]{narayan2018don}
Shashi Narayan, Shay~B. Cohen, and Mirella Lapata. 2018.
\newblock \href {https://doi.org/10.18653/v1/D18-1206} {Don{'}t give me the
  details, just the summary! topic-aware convolutional neural networks for
  extreme summarization}.
\newblock In \emph{Proceedings of the 2018 Conference on Empirical Methods in
  Natural Language Processing}, pages 1797--1807, Brussels, Belgium.
  Association for Computational Linguistics.

\bibitem[{Novikova et~al.(2017)Novikova, Du{\v{s}}ek, and
  Rieser}]{novikova2017e2e}
Jekaterina Novikova, Ond{\v{r}}ej Du{\v{s}}ek, and Verena Rieser. 2017.
\newblock \href {https://doi.org/10.18653/v1/W17-5525} {The {E}2{E} dataset:
  New challenges for end-to-end generation}.
\newblock In \emph{Proceedings of the 18th Annual {SIG}dial Meeting on
  Discourse and Dialogue}, pages 201--206, Saarbr{\"u}cken, Germany.
  Association for Computational Linguistics.

\bibitem[{Opitz and Frank(2020)}]{opitz2020towards}
Juri Opitz and Anette Frank. 2020.
\newblock Towards a decomposable metric for explainable evaluation of text
  generation from amr.
\newblock \emph{arXiv preprint arXiv:2008.08896}.

\bibitem[{Papineni et~al.(2002)Papineni, Roukos, Ward, and
  Zhu}]{papineni2002bleu}
Kishore Papineni, Salim Roukos, Todd Ward, and Wei-Jing Zhu. 2002.
\newblock \href {https://doi.org/10.3115/1073083.1073135} {{B}leu: a method for
  automatic evaluation of machine translation}.
\newblock In \emph{Proceedings of the 40th Annual Meeting of the Association
  for Computational Linguistics}, pages 311--318, Philadelphia, Pennsylvania,
  USA. Association for Computational Linguistics.

\bibitem[{Parikh et~al.(2020)Parikh, Wang, Gehrmann, Faruqui, Dhingra, Yang,
  and Das}]{parikh2020totto}
Ankur Parikh, Xuezhi Wang, Sebastian Gehrmann, Manaal Faruqui, Bhuwan Dhingra,
  Diyi Yang, and Dipanjan Das. 2020.
\newblock \href {https://doi.org/10.18653/v1/2020.emnlp-main.89} {{ToTTo}: A
  controlled table-to-text generation dataset}.
\newblock In \emph{Proceedings of the 2020 Conference on Empirical Methods in
  Natural Language Processing (EMNLP)}, pages 1173--1186, Online. Association
  for Computational Linguistics.

\bibitem[{Perez-Beltrachini and Gardent(2017)}]{perez2017analysing}
Laura Perez-Beltrachini and Claire Gardent. 2017.
\newblock \href {https://doi.org/10.18653/v1/W17-3537} {Analysing data-to-text
  generation benchmarks}.
\newblock In \emph{Proceedings of the 10th International Conference on Natural
  Language Generation}, pages 238--242, Santiago de Compostela, Spain.
  Association for Computational Linguistics.

\bibitem[{Petroni et~al.(2020)Petroni, Piktus, Fan, Lewis, Yazdani, Cao,
  Thorne, Jernite, Plachouras, Rockt{\"{a}}schel, and Riedel}]{petroni2020kilt}
Fabio Petroni, Aleksandra Piktus, Angela Fan, Patrick S.~H. Lewis, Majid
  Yazdani, Nicola~De Cao, James Thorne, Yacine Jernite, Vassilis Plachouras,
  Tim Rockt{\"{a}}schel, and Sebastian Riedel. 2020.
\newblock \href {http://arxiv.org/abs/2009.02252} {{KILT:} a benchmark for
  knowledge intensive language tasks}.
\newblock \emph{CoRR}, abs/2009.02252.

\bibitem[{Ponti et~al.(2020)Ponti, Glava{\v{s}}, Majewska, Liu, Vuli{\'c}, and
  Korhonen}]{ponti2020xcopa}
Edoardo~Maria Ponti, Goran Glava{\v{s}}, Olga Majewska, Qianchu Liu, Ivan
  Vuli{\'c}, and Anna Korhonen. 2020.
\newblock \href {https://doi.org/10.18653/v1/2020.emnlp-main.185} {{XCOPA}: A
  multilingual dataset for causal commonsense reasoning}.
\newblock In \emph{Proceedings of the 2020 Conference on Empirical Methods in
  Natural Language Processing (EMNLP)}, pages 2362--2376, Online. Association
  for Computational Linguistics.

\bibitem[{Potts et~al.(2020)Potts, Wu, Geiger, and Kiela}]{potts2020dynasent}
Christopher Potts, Zhengxuan Wu, Atticus Geiger, and Douwe Kiela. 2020.
\newblock \href {http://arxiv.org/abs/2012.15349} {Dynasent: {A} dynamic
  benchmark for sentiment analysis}.
\newblock \emph{CoRR}, abs/2012.15349.

\bibitem[{Puduppully et~al.(2019)Puduppully, Dong, and
  Lapata}]{puduppully-etal-2019-data}
Ratish Puduppully, Li~Dong, and Mirella Lapata. 2019.
\newblock \href {https://doi.org/10.18653/v1/P19-1195} {Data-to-text generation
  with entity modeling}.
\newblock In \emph{Proceedings of the 57th Annual Meeting of the Association
  for Computational Linguistics}, pages 2023--2035, Florence, Italy.
  Association for Computational Linguistics.

\bibitem[{Radev et~al.(2020)Radev, Zhang, Rau, Sivaprasad, Hsieh, Rajani, Tang,
  Vyas, Verma, Krishna, Liu, Irwanto, Pan, Rahman, Zaidi, Mutuma, Tarabar,
  Gupta, Yu, Tan, Lin, Xiong, and Socher}]{radev2020dart}
Dragomir~R. Radev, Rui Zhang, Amrit Rau, Abhinand Sivaprasad, Chiachun Hsieh,
  Nazneen~Fatema Rajani, Xiangru Tang, Aadit Vyas, Neha Verma, Pranav Krishna,
  Yangxiaokang Liu, Nadia Irwanto, Jessica Pan, Faiaz Rahman, Ahmad Zaidi,
  Murori Mutuma, Yasin Tarabar, Ankit Gupta, Tao Yu, Yi~Chern Tan, Xi~Victoria
  Lin, Caiming Xiong, and Richard Socher. 2020.
\newblock \href {http://arxiv.org/abs/2007.02871} {{DART:} open-domain
  structured data record to text generation}.
\newblock \emph{CoRR}, abs/2007.02871.

\bibitem[{Raffel et~al.(2020)Raffel, Shazeer, Roberts, Lee, Narang, Matena,
  Zhou, Li, and Liu}]{raffel2019exploring}
Colin Raffel, Noam Shazeer, Adam Roberts, Katherine Lee, Sharan Narang, Michael
  Matena, Yanqi Zhou, Wei Li, and Peter~J. Liu. 2020.
\newblock \href {http://jmlr.org/papers/v21/20-074.html} {Exploring the limits
  of transfer learning with a unified text-to-text transformer}.
\newblock \emph{J. Mach. Learn. Res.}, 21:140:1--140:67.

\bibitem[{Rao and Tetreault(2018)}]{rao2018dear}
Sudha Rao and Joel Tetreault. 2018.
\newblock \href {https://doi.org/10.18653/v1/N18-1012} {Dear sir or madam, may
  {I} introduce the {GYAFC} dataset: Corpus, benchmarks and metrics for
  formality style transfer}.
\newblock In \emph{Proceedings of the 2018 Conference of the North {A}merican
  Chapter of the Association for Computational Linguistics: Human Language
  Technologies, Volume 1 (Long Papers)}, pages 129--140, New Orleans,
  Louisiana. Association for Computational Linguistics.

\bibitem[{Rastogi et~al.(2020)Rastogi, Zang, Sunkara, Gupta, and
  Khaitan}]{rastogi2019towards}
Abhinav Rastogi, Xiaoxue Zang, Srinivas Sunkara, Raghav Gupta, and Pranav
  Khaitan. 2020.
\newblock \href {https://aaai.org/ojs/index.php/AAAI/article/view/6394}
  {Towards scalable multi-domain conversational agents: The schema-guided
  dialogue dataset}.
\newblock In \emph{The Thirty-Fourth {AAAI} Conference on Artificial
  Intelligence, {AAAI} 2020, The Thirty-Second Innovative Applications of
  Artificial Intelligence Conference, {IAAI} 2020, The Tenth {AAAI} Symposium
  on Educational Advances in Artificial Intelligence, {EAAI} 2020, New York,
  NY, USA, February 7-12, 2020}, pages 8689--8696. {AAAI} Press.

\bibitem[{Reddy et~al.(2019)Reddy, Chen, and Manning}]{reddy-etal-2019-coqa}
Siva Reddy, Danqi Chen, and Christopher~D. Manning. 2019.
\newblock \href {https://doi.org/10.1162/tacl_a_00266} {{C}o{QA}: A
  conversational question answering challenge}.
\newblock \emph{Transactions of the Association for Computational Linguistics},
  7:249--266.

\bibitem[{Reiter(2018)}]{reiter2018structured}
Ehud Reiter. 2018.
\newblock \href {https://doi.org/10.1162/coli\_a\_00322} {A structured review
  of the validity of {BLEU}}.
\newblock \emph{Comput. Linguistics}, 44(3).

\bibitem[{Reiter and Dale(2000)}]{reiter2000building}
Ehud Reiter and Robert Dale. 2000.
\newblock \emph{Building natural language generation systems}.
\newblock Cambridge university press.

\bibitem[{Ribeiro et~al.(2020)Ribeiro, Wu, Guestrin, and
  Singh}]{ribeiro-etal-2020-beyond}
Marco~Tulio Ribeiro, Tongshuang Wu, Carlos Guestrin, and Sameer Singh. 2020.
\newblock \href {https://doi.org/10.18653/v1/2020.acl-main.442} {Beyond
  accuracy: Behavioral testing of {NLP} models with {C}heck{L}ist}.
\newblock In \emph{Proceedings of the 58th Annual Meeting of the Association
  for Computational Linguistics}, pages 4902--4912, Online. Association for
  Computational Linguistics.

\bibitem[{Schlegel et~al.(2020)Schlegel, Nenadic, and
  Batista{-}Navarro}]{schlegel2020beyond}
Viktor Schlegel, Goran Nenadic, and Riza Batista{-}Navarro. 2020.
\newblock \href {http://arxiv.org/abs/2005.14709} {Beyond leaderboards: {A}
  survey of methods for revealing weaknesses in natural language inference data
  and models}.
\newblock \emph{CoRR}, abs/2005.14709.

\bibitem[{Scialom et~al.(2020)Scialom, Dray, Lamprier, Piwowarski, and
  Staiano}]{scialom2020mlsum}
Thomas Scialom, Paul-Alexis Dray, Sylvain Lamprier, Benjamin Piwowarski, and
  Jacopo Staiano. 2020.
\newblock \href {https://doi.org/10.18653/v1/2020.emnlp-main.647} {{MLSUM}: The
  multilingual summarization corpus}.
\newblock In \emph{Proceedings of the 2020 Conference on Empirical Methods in
  Natural Language Processing (EMNLP)}, pages 8051--8067, Online. Association
  for Computational Linguistics.

\bibitem[{Scialom et~al.(2021)Scialom, Dray, Patrick, Sylvain, Benjamin,
  Jacopo, and Alex}]{scialom2020SAFEval}
Thomas Scialom, Paul-Alexis Dray, Gallinari Patrick, Lamprier Sylvain,
  Piwowarski Benjamin, Staiano Jacopo, and Wang Alex. 2021.
\newblock Safeval: Summarization asks for fact-based evaluation.
\newblock \emph{arXiv preprint arXiv:2103.12693}.

\bibitem[{Scialom et~al.(2019)Scialom, Lamprier, Piwowarski, and
  Staiano}]{scialom-2019-answers}
Thomas Scialom, Sylvain Lamprier, Benjamin Piwowarski, and Jacopo Staiano.
  2019.
\newblock \href {https://doi.org/10.18653/v1/D19-1320} {Answers unite!
  unsupervised metrics for reinforced summarization models}.
\newblock In \emph{Proceedings of the 2019 Conference on Empirical Methods in
  Natural Language Processing and the 9th International Joint Conference on
  Natural Language Processing (EMNLP-IJCNLP)}, pages 3246--3256, Hong Kong,
  China. Association for Computational Linguistics.

\bibitem[{Sellam et~al.(2020)Sellam, Das, and Parikh}]{sellam-2020-bleurt}
Thibault Sellam, Dipanjan Das, and Ankur Parikh. 2020.
\newblock \href {https://doi.org/10.18653/v1/2020.acl-main.704} {{BLEURT}:
  Learning robust metrics for text generation}.
\newblock In \emph{Proceedings of the 58th Annual Meeting of the Association
  for Computational Linguistics}, pages 7881--7892, Online. Association for
  Computational Linguistics.

\bibitem[{Shannon and Weaver(1963)}]{shannon2001mathematical}
Claude~E Shannon and Warren Weaver. 1963.
\newblock A mathematical theory of communication.

\bibitem[{Sharma et~al.(2019)Sharma, Li, and Wang}]{sharma2019bigpatent}
Eva Sharma, Chen Li, and Lu~Wang. 2019.
\newblock \href {https://doi.org/10.18653/v1/P19-1212} {{BIGPATENT}: A
  large-scale dataset for abstractive and coherent summarization}.
\newblock In \emph{Proceedings of the 57th Annual Meeting of the Association
  for Computational Linguistics}, pages 2204--2213, Florence, Italy.
  Association for Computational Linguistics.

\bibitem[{Shimorina and Belz(2021)}]{shimorina2021human}
Anastasia Shimorina and Anya Belz. 2021.
\newblock The human evaluation datasheet 1.0: A template for recording details
  of human evaluation experiments in nlp.
\newblock \emph{arXiv preprint arXiv:2103.09710}.

\bibitem[{Shukla et~al.(2019)Shukla, Elmadjian, Sharan, Kulkarni, Turk, and
  Wang}]{shukla-etal-2019-ask}
Pushkar Shukla, Carlos Elmadjian, Richika Sharan, Vivek Kulkarni, Matthew Turk,
  and William~Yang Wang. 2019.
\newblock \href {https://doi.org/10.18653/v1/P19-1646} {What should {I} ask?
  using conversationally informative rewards for goal-oriented visual dialog.}
\newblock In \emph{Proceedings of the 57th Annual Meeting of the Association
  for Computational Linguistics}, pages 6442--6451, Florence, Italy.
  Association for Computational Linguistics.

\bibitem[{Stanojevi{\'c} et~al.(2015)Stanojevi{\'c}, Kamran, Koehn, and
  Bojar}]{stanojevic2015results}
Milo{\v{s}} Stanojevi{\'c}, Amir Kamran, Philipp Koehn, and Ond{\v{r}}ej Bojar.
  2015.
\newblock \href {https://doi.org/10.18653/v1/W15-3031} {Results of the {WMT}15
  metrics shared task}.
\newblock In \emph{Proceedings of the Tenth Workshop on Statistical Machine
  Translation}, pages 256--273, Lisbon, Portugal. Association for Computational
  Linguistics.

\bibitem[{Sun et~al.(2019)Sun, Shapira, Dagan, and Nenkova}]{sun2019compare}
Simeng Sun, Ori Shapira, Ido Dagan, and Ani Nenkova. 2019.
\newblock \href {https://doi.org/10.18653/v1/W19-2303} {How to compare
  summarizers without target length? pitfalls, solutions and re-examination of
  the neural summarization literature}.
\newblock In \emph{Proceedings of the Workshop on Methods for Optimizing and
  Evaluating Neural Language Generation}, pages 21--29, Minneapolis, Minnesota.
  Association for Computational Linguistics.

\bibitem[{Toutanova et~al.(2016)Toutanova, Brockett, Tran, and
  Amershi}]{toutanova-etal-2016-dataset}
Kristina Toutanova, Chris Brockett, Ke~M. Tran, and Saleema Amershi. 2016.
\newblock \href {https://doi.org/10.18653/v1/D16-1033} {A dataset and
  evaluation metrics for abstractive compression of sentences and short
  paragraphs}.
\newblock In \emph{Proceedings of the 2016 Conference on Empirical Methods in
  Natural Language Processing}, pages 340--350, Austin, Texas. Association for
  Computational Linguistics.

\bibitem[{Wang et~al.(2020)Wang, Cho, and Lewis}]{wang-2020-asking}
Alex Wang, Kyunghyun Cho, and Mike Lewis. 2020.
\newblock \href {https://doi.org/10.18653/v1/2020.acl-main.450} {Asking and
  answering questions to evaluate the factual consistency of summaries}.
\newblock In \emph{Proceedings of the 58th Annual Meeting of the Association
  for Computational Linguistics}, pages 5008--5020, Online. Association for
  Computational Linguistics.

\bibitem[{Wang et~al.(2019{\natexlab{a}})Wang, Pruksachatkun, Nangia, Singh,
  Michael, Hill, Levy, and Bowman}]{wang2019superglue}
Alex Wang, Yada Pruksachatkun, Nikita Nangia, Amanpreet Singh, Julian Michael,
  Felix Hill, Omer Levy, and Samuel~R. Bowman. 2019{\natexlab{a}}.
\newblock \href
  {https://proceedings.neurips.cc/paper/2019/hash/4496bf24afe7fab6f046bf4923da8de6-Abstract.html}
  {Superglue: {A} stickier benchmark for general-purpose language understanding
  systems}.
\newblock In \emph{Advances in Neural Information Processing Systems 32: Annual
  Conference on Neural Information Processing Systems 2019, NeurIPS 2019,
  December 8-14, 2019, Vancouver, BC, Canada}, pages 3261--3275.

\bibitem[{Wang et~al.(2019{\natexlab{b}})Wang, Singh, Michael, Hill, Levy, and
  Bowman}]{wang2018glue}
Alex Wang, Amanpreet Singh, Julian Michael, Felix Hill, Omer Levy, and
  Samuel~R. Bowman. 2019{\natexlab{b}}.
\newblock \href {https://openreview.net/forum?id=rJ4km2R5t7} {{GLUE}: A
  multi-task benchmark and analysis platform for natural language
  understanding}.
\newblock In \emph{International Conference on Learning Representations}.

\bibitem[{Wiseman et~al.(2017)Wiseman, Shieber, and
  Rush}]{wiseman-etal-2017-challenges}
Sam Wiseman, Stuart Shieber, and Alexander Rush. 2017.
\newblock \href {https://doi.org/10.18653/v1/D17-1239} {Challenges in
  data-to-document generation}.
\newblock In \emph{Proceedings of the 2017 Conference on Empirical Methods in
  Natural Language Processing}, pages 2253--2263, Copenhagen, Denmark.
  Association for Computational Linguistics.

\bibitem[{Xie et~al.(2020)Xie, Dai, Hovy, Luong, and Le}]{xie2020unsupervised}
Qizhe Xie, Zihang Dai, Eduard Hovy, Thang Luong, and Quoc Le. 2020.
\newblock Unsupervised data augmentation for consistency training.
\newblock \emph{Advances in Neural Information Processing Systems}, 33.

\bibitem[{Xu et~al.(2016)Xu, Napoles, Pavlick, Chen, and
  Callison-Burch}]{xu-etal-2016-optimizing}
Wei Xu, Courtney Napoles, Ellie Pavlick, Quanze Chen, and Chris Callison-Burch.
  2016.
\newblock \href {https://doi.org/10.1162/tacl_a_00107} {Optimizing statistical
  machine translation for text simplification}.
\newblock \emph{Transactions of the Association for Computational Linguistics},
  4:401--415.

\bibitem[{Xue et~al.(2020)Xue, Constant, Roberts, Kale, Al{-}Rfou, Siddhant,
  Barua, and Raffel}]{xue2020mt5}
Linting Xue, Noah Constant, Adam Roberts, Mihir Kale, Rami Al{-}Rfou, Aditya
  Siddhant, Aditya Barua, and Colin Raffel. 2020.
\newblock \href {http://arxiv.org/abs/2010.11934} {mt5: {A} massively
  multilingual pre-trained text-to-text transformer}.
\newblock \emph{CoRR}, abs/2010.11934.

\bibitem[{Zang et~al.(2020)Zang, Rastogi, Sunkara, Gupta, Zhang, and
  Chen}]{zang2020multiwoz}
Xiaoxue Zang, Abhinav Rastogi, Srinivas Sunkara, Raghav Gupta, Jianguo Zhang,
  and Jindong Chen. 2020.
\newblock \href {https://doi.org/10.18653/v1/2020.nlp4convai-1.13}
  {{M}ulti{WOZ} 2.2 : A dialogue dataset with additional annotation corrections
  and state tracking baselines}.
\newblock In \emph{Proceedings of the 2nd Workshop on Natural Language
  Processing for Conversational AI}, pages 109--117, Online. Association for
  Computational Linguistics.

\bibitem[{Zhang et~al.(2020{\natexlab{a}})Zhang, Zhao, Saleh, and
  Liu}]{zhang2020pegasus}
Jingqing Zhang, Yao Zhao, Mohammad Saleh, and Peter~J. Liu. 2020{\natexlab{a}}.
\newblock \href {http://proceedings.mlr.press/v119/zhang20ae.html} {{PEGASUS:}
  pre-training with extracted gap-sentences for abstractive summarization}.
\newblock In \emph{Proceedings of the 37th International Conference on Machine
  Learning, {ICML} 2020, 13-18 July 2020, Virtual Event}, volume 119 of
  \emph{Proceedings of Machine Learning Research}, pages 11328--11339. {PMLR}.

\bibitem[{Zhang et~al.(2018)Zhang, Dinan, Urbanek, Szlam, Kiela, and
  Weston}]{zhang2018personalizing}
Saizheng Zhang, Emily Dinan, Jack Urbanek, Arthur Szlam, Douwe Kiela, and Jason
  Weston. 2018.
\newblock \href {https://doi.org/10.18653/v1/P18-1205} {Personalizing dialogue
  agents: {I} have a dog, do you have pets too?}
\newblock In \emph{Proceedings of the 56th Annual Meeting of the Association
  for Computational Linguistics (Volume 1: Long Papers)}, pages 2204--2213,
  Melbourne, Australia. Association for Computational Linguistics.

\bibitem[{Zhang et~al.(2020{\natexlab{b}})Zhang, Kishore, Wu, Weinberger, and
  Artzi}]{zhang2019bertscore}
Tianyi Zhang, Varsha Kishore, Felix Wu, Kilian~Q. Weinberger, and Yoav Artzi.
  2020{\natexlab{b}}.
\newblock \href {https://openreview.net/forum?id=SkeHuCVFDr} {Bertscore:
  Evaluating text generation with {BERT}}.
\newblock In \emph{8th International Conference on Learning Representations,
  {ICLR} 2020, Addis Ababa, Ethiopia, April 26-30, 2020}. OpenReview.net.

\bibitem[{Zhang and Lapata(2014)}]{zhang-lapata-2014-chinese}
Xingxing Zhang and Mirella Lapata. 2014.
\newblock \href {https://doi.org/10.3115/v1/D14-1074} {{C}hinese poetry
  generation with recurrent neural networks}.
\newblock In \emph{Proceedings of the 2014 Conference on Empirical Methods in
  Natural Language Processing ({EMNLP})}, pages 670--680, Doha, Qatar.
  Association for Computational Linguistics.

\bibitem[{Zhang et~al.(2020{\natexlab{c}})Zhang, Sun, Galley, Chen, Brockett,
  Gao, Gao, Liu, and Dolan}]{zhang2019dialogpt}
Yizhe Zhang, Siqi Sun, Michel Galley, Yen-Chun Chen, Chris Brockett, Xiang Gao,
  Jianfeng Gao, Jingjing Liu, and Bill Dolan. 2020{\natexlab{c}}.
\newblock \href {https://doi.org/10.18653/v1/2020.acl-demos.30} {{DIALOGPT} :
  Large-scale generative pre-training for conversational response generation}.
\newblock In \emph{Proceedings of the 58th Annual Meeting of the Association
  for Computational Linguistics: System Demonstrations}, pages 270--278,
  Online. Association for Computational Linguistics.

\end{thebibliography}
\bibliographystyle{acl_natbib}

\appendix 

\section{Task Suggestion Categories}
\label{app:suggestion}
Participants were required to provide information for the following categories when suggesting a dataset for \GEM.

\begin{enumerate}[itemsep=0mm]
    \item Dataset Name
    \item Reference
    \item High-level Task, e.g., data-to-text, or summarization
    \item Short Description
    \item Challenges, e.g., entity tracking/generation, referring expression generation, surface realization, content selection
    \item Communicative goal, e.g., provide specific information, or entertainment, or accomplish a task
    \item Dataset Domain, e.g., Wikipedia, or news articles, Reddit chat, etc)
    \item Language(s)
    \item Language locale (if known), e.g., en-US, es-MX
    \item Input modality, e.g., text, graph, table, images
    \item Input length
    \item Output length
    \item Output form, e.g., monologue, dialog
    \item \# Examples in dataset	Test split, e.g., i.i.d., or non-overlap dimension
    \item \# References per example
    \item Data Quality / potential Issues, e.g., noisy, clean, biased, code-mixing (different languages/writing systems), (over)-normalization
    \item License
    \item Evaluation strategies (in original paper / papers that use dataset)
    \item Why should we use this dataset?
\end{enumerate}

\section{Considered datasets}
\label{app:considered}

The following datasets were proposed to be included in \GEM.

\begin{enumerate}[itemsep=0mm]
    \item Alex Context NLG~\citep{duvsek28context,dusek2016context}
    \item AmbigQA/AmbigNQ~\citep{min2020ambigqa}
    \item Bangla Natural Language Image to Text~\citep{jishan2020bangla}
    \item Big Patent~\citep{sharma2019bigpatent}
    \item Chinese Poetry~\citep{zhang-lapata-2014-chinese}
    \item CommonGen~\citep{lin2019commongen}
    \item CoQA~\citep{reddy-etal-2019-coqa}
    \item Czech Restaurant Data~\citep{duvsek2019neural}
    \item DART~\citep{radev2020dart}
    \item E2E (cleaned)~\citep{novikova2017e2e,duvsek2019semantic}
    \item ELI5~\citep{fan-etal-2019-eli5}
    \item Hindi Poetry~\footnote{https://www.kaggle.com/shishu1421/hindi-poetry-dataset}
    \item LCSTS~\citep{hu-etal-2015-lcsts}
    \item Mimic and Rephrase~\citep{dieter-etal-2019-mimic}
    \item MLSUM~\citep{scialom2020mlsum}
    \item MSR Abstractive Text Compression~\citep{toutanova-etal-2016-dataset}
    \item MultiWOZ 2.2~\citep{zang2020multiwoz}
    \item NarrativeQA~\citep{kovcisky2018narrativeqa}
    \item PersonaChat~\citep{zhang2018personalizing}
    \item PubMed, Arxiv~\citep{kedzie2018content,cohan2018discourse}
    \item ROTOWIRE/MLB~\citep{wiseman-etal-2017-challenges,puduppully-etal-2019-data}
    \item Schema-Guided Dialogue~\citep{rastogi2019towards}
    \item SQUAD Question Generation~\citep{du2017learning}
    \item SR'11, SR'18, SR'19~\citep{belz-etal-2011-first,mille2018first,mille2019proceedings}
    \item ToTTo~\citep{parikh2020totto}
    \item Ubuntu Dialogue Generation~\citep{lowe-etal-2015-ubuntu}
    \item Visual Question Generation~\citep{shukla-etal-2019-ask,li2018visual}
    \item WebNLG~\citep{gardent2017webnlg}
    \item WikiAuto + Turk/ASSET~\citep{jiang-etal-2020-neural,xu-etal-2016-optimizing,alva-manchego-etal-2020-asset}
    \item WikiBio~\citep{lebret2016neural}
    \item WikiSum~\citep{liu2018generating}
    \item Wizard of Wikipedia~\citep{dinan2018wizard}
    \item Writing Prompts~\citep{fan-etal-2018-hierarchical}
    \item XSum~\citep{narayan2018don}
    \item WikiLingua~\citep{ladhak2020wikilingua}
\end{enumerate}

\section{Task and Criteria Selection Survey}
\label{app:survey}

As part of our selection process, we queried all \GEM{} members about the utility of tasks and selection criteria. The questions below were included in the survey. 

\begin{itemize}[itemsep=0mm]
    \item For each suggested task, ``Should this task be included in GEM?'' on a 5-point Likert scale (1 being \textit{strongly against}, and 5 \textit{strongly in favor}). 
    \item We should exclude tasks that are the focus of a shared task in 2021. [yes/no]
    \item We should exclude tasks that were the focus of a shared task since 2020. [yes/no]
    \item We should exclude tasks that were ever part of a shared task. [yes/no]
    \item We should exclude datasets that require paid-for licenses (e.g., LDC or ELRA). [yes/no]
    \item We should exclude datasets that are not freely available for download. [yes/no]
    \item We should exclude tasks that require encoding anything but text (e.g., images or graphs). [yes/no]
    \item We should include \# tasks in GEM. [10 points ranging from 2 to 20]
    \item X\% of the tasks should feature non-English language(s). [10 points ranging from 10 to 100\%]
    \item Diversity of tasks is more important than focus on an NLG task (by including multiple datasets for the same task). [10 points from \textit{Diversity is more important} to \textit{Focus is more important}]
    \item We should include noisy and clean datasets. [10 points from \textit{only noisy} to \textit{only clean}]
    \item We should include low- and high-resource datasets. [10 points from \textit{only low-resource} to \textit{only high-resource}]
    \item We should prefer tasks with non-iid test sets or specific challenge sets. [5-Likert scale from \textit{not important} to \textit{very important}]
    \item We should prefer tasks with test sets with multiple references. [5-Likert scale from \textit{not important} to \textit{very important}]
    \item If we include an NLG task (e.g., simplification or data2text), we need multiple datasets for that task. [5-Likert scale from \textit{not important} to \textit{very important}]
    \item We should include a set of tasks with no clear evaluation strategy. [5-Likert scale from \textit{not important} to \textit{very important}]
    \item We should focus on tasks with reliable automatic metrics. [5-Likert scale from \textit{not important} to \textit{very important}]
    
\end{itemize}

\end{document}